\documentclass[twocolumn, DIV25, 9pt]{scrartcl}
\date{}
\pdfoutput=1

\usepackage{palatino}
\usepackage{graphicx}
\usepackage{subfigure}
\usepackage{float}
\usepackage{lettrine}
\usepackage{compactbib}
\usepackage{cite}
\usepackage{url}
\usepackage{float}
\usepackage{amsmath}
\usepackage[noend]{algpseudocode}%

\newif\iftodos

\todostrue

\usepackage[usenames,dvipsnames]{color} %
\iftodos

\newcommand{\todo}[1]{\textcolor{red}{[#1]}} %
\newcommand{\done}[1]{\textcolor{Emerald}{[#1]}} %
\newcommand{\comment}[1]{\textcolor{blue}{[#1]}} %
\newcommand{\jmccomment}[1]{\textcolor{magenta}{[#1]}} %
\newcommand{\lowpriority}[1]{\textcolor{green}{[#1]}} %
\newcommand{\todoOfficialVersion}[1]{} %
\else
\newcommand{\todo}[1]{} %
\newcommand{\done}[1]{} %
\newcommand{\comment}[1]{} %
\newcommand{\jmccomment}[1]{} %
\newcommand{\lowpriority}[1]{} %
\fi
\usepackage[labelfont={bf},small,sf]{caption}
\DeclareCaptionLabelSeparator{pp}{.}
\setcapindent{0cm}
\DeclareCaptionType{Supplementary figure}[Fig.]
\DeclareCaptionLabelFormat{bf-parens}{#1 S#2.}
\makeatletter
\renewcommand{\@biblabel}[1]{\quad#1.}
\makeatother

\RequirePackage{fancyhdr}
\title{\flushleft{\fontsize{21}{12}\selectfont \textsf{\textbf{Illuminating search spaces by mapping elites}}\\
  \textbf{\large{\textsf{
        Jean-Baptiste Mouret$\mathsf{^{1}}$ and 
        Jeff Clune$\mathsf{^{2}}$}}}\\
  ~\\
         \small{
           $\mathsf{^1}$Universit\'e Pierre et Marie Curie-Paris 6, CNRS UMR 7222, France\\
           $\mathsf{^2}$University of Wyoming, USA\\
}}
\noindent{\normalsize \textsf{Preprint -- \today}}
\vspace*{-1.5cm}
}
\makeatletter
\def\@cite#1#2{$^{\mbox{\scriptsize #1\if@tempswa , #2\fi}}$}
\makeatother

\begin{document}

\maketitle

\thispagestyle{fancy}
\pagestyle{fancy}
\bibliographystyle{plain}

\begin{abstract}
\sffamily \bfseries 

Nearly all science and engineering fields use \emph{search algorithms}, which automatically explore a search space to find high-performing solutions: chemists search through the space of molecules to discover new drugs; engineers search for stronger, cheaper, safer designs, scientists search for models that best explain data, etc. The goal of search algorithms has traditionally been to return the single highest-performing solution in a search space. Here we describe a new, fundamentally different type of algorithm that is more useful because it provides a holistic view of how high-performing solutions are distributed throughout a search space. It creates a map of high-performing solutions at each point in a space defined by dimensions of variation that a user gets to choose. This Multi-dimensional Archive of Phenotypic Elites (MAP-Elites) algorithm illuminates search spaces, allowing researchers to understand how interesting attributes of solutions combine to affect performance, either positively or, equally of interest, negatively. For example, a drug company may wish to understand how performance changes as the size of molecules and their cost-to-produce vary.  MAP-Elites produces a large diversity of high-performing, yet qualitatively different solutions, which can be more helpful than a single, high-performing solution. Interestingly, because MAP-Elites explores more of the search space, it also tends to find a better overall solution than state-of-the-art search algorithms. We demonstrate the benefits of this new algorithm in three different problem domains ranging from producing modular neural networks to designing simulated and real soft robots. Because MAP-Elites (1) illuminates the relationship between performance and dimensions of interest in solutions, (2) returns a set of high-performing, yet diverse solutions, and (3) improves the state-of-the-art for finding a single, best solution, it will catalyze advances throughout all science and engineering fields.
\end{abstract}

\vspace{3ex}
\emph{
Author's Note: This paper is a preliminary draft of a paper that introduces the MAP-Elites algorithm and explores its capabilities. Normally we would not post such an early draft with only preliminary experimental data, but many people in the community have heard of MAP-Elites, are using it in their own papers, and have asked us for a paper that describes it so that they can cite it, to help them implement MAP-Elites, and that describes the experiments we have already conducted with it. We thus want to share both the details of this algorithm and what we have learned about it from our preliminary experiments. All of the experiments in this paper will be redone before the final version of the paper is published, and the data are thus subject to change.} 

\section{Background and Motivation}
\vspace{2ex}
Every field of science and engineering makes use of search algorithms, also known as optimization algorithms, which seek to automatically find a high-quality solution or set of high-quality solutions amongst a large space of possible solutions\cite{russell1995artificial,floreano2008bio}. Such algorithms often find solutions that outperform those designed by human engineers\cite{koza2003genetic}: they have designed antennas that NASA flew to space\cite{hornby2011computer}, found patentable electronic circuit designs\cite{koza2003genetic}, automated scientific discovery\cite{schmidt2009distilling}, and created artificial intelligence for robots\cite{cully2015robots,clune2011performance, cheney2013unshackling,hornby2005autonomous,doncieux2015evolutionary, yosinski2011gaits, lee2013evolving, clune2009evolving, cully2015evolving, lipson2000automatic}. Because of their widespread use, improving search algorithms provides substantial benefits for society.

Most search algorithms focus on finding one or a small set of high-quality solutions in a search space. What constitutes high-quality is determined by the user, who specifies one or a few objectives that the solution should score high on. For example, a user may want solutions that are high-performing and low-cost, where each of those desiderata is quantifiably measured either by an equation or simulator. Traditional search algorithms include hill climbing, simulated annealing, evolutionary algorithms, gradient ascent/descent, Bayesian optimization, and multi-objective optimization algorithms\cite{russell1995artificial,floreano2008bio}. The latter return a set of solutions that represent the best tradeoffs between objectives\cite{deb2001multi}. 

\begin{figure}
\centering
\includegraphics[width=\linewidth]{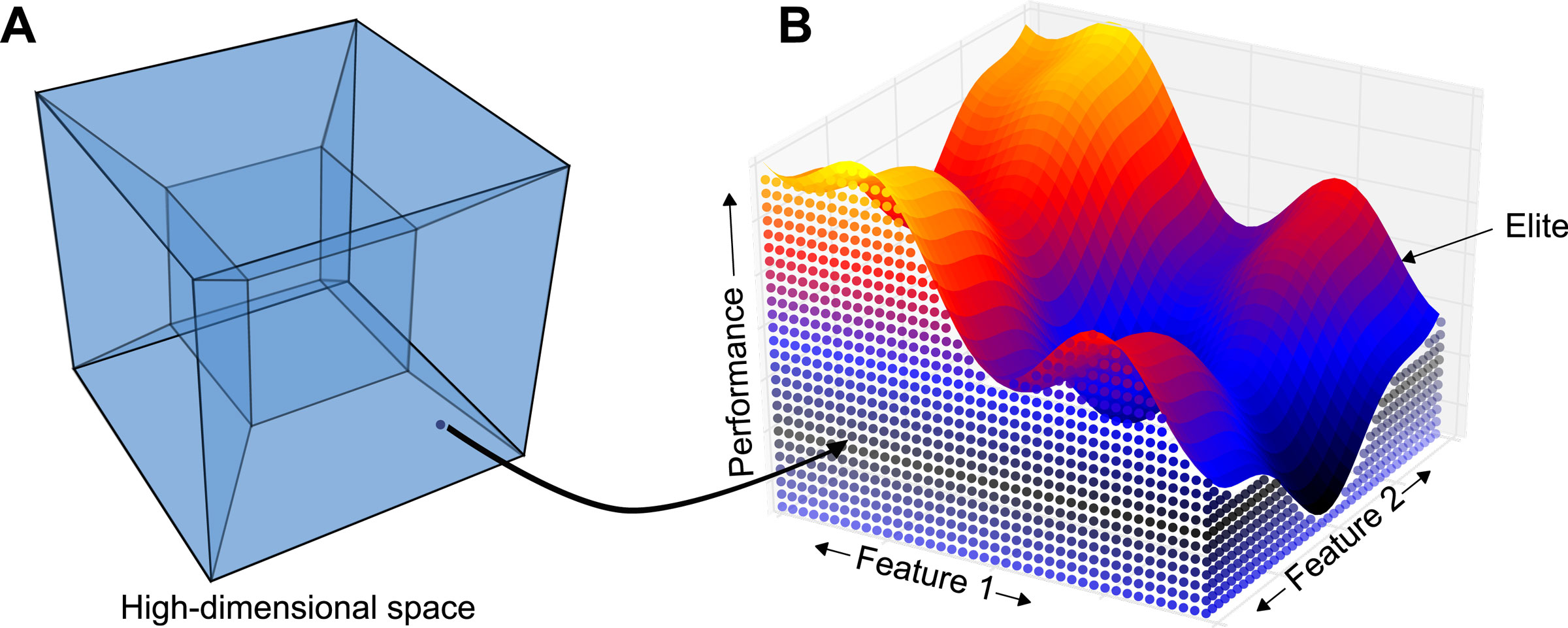}
\caption{The MAP-Elites algorithm searches in a high-dimensional space to find the highest-performing solution at each point in a low-dimensional feature space, where the user gets to choose dimensions of variation of interest that define the low dimensional space. We call this type of algorithm an ``illumination algorithm'', because it illuminates the fitness potential of each area of the feature space, including tradeoffs between performance and the features of interest. For example, MAP-Elites could search in the space of all possible robot designs (a very high dimensional space) to find the fastest robot (a performance criterion) for each combination of height and weight.}
\label{conceptFig}
\end{figure}

A subset of optimization problems are challenging because they require searching for optima in a function or system that is either non-differentiable or cannot be expressed mathematically, typically because a physical system or a complex simulation is required. Such problems require ``black box" optimization algorithms, which search for high-performing solutions armed only with the ability to determine the performance of a solution, but without access to the evaluation function that determines that performance. On such problems, one cannot use optimization methods that require calculating the gradient of the function, such as gradient ascent/descent. 
 
A notorious challenge in black box optimization is the presence of local optima (also called local minima)\cite{russell1995artificial,floreano2008bio}.
A problem with most search algorithms of this class is that they try to follow a path that will lead to the best global solution by relying on the heuristic that random changes to good solutions lead to better solutions. This approach does not work for highly deceptive problems, however, because in such problems one has to cross low-performing valleys to find the global optima, or even just to find better optima\cite{floreano2008bio}.

Because evolutionary algorithms are one of the most successful families of black-box search algorithms\cite{koza2003genetic, floreano2008bio}, and because the work we build on comes from that community, here we adopt the language and metaphors of evolutionary computation. In that parlance, a solution is an \emph{organism} or \emph{phenotype} or \emph{individual}, the organism is described by a \emph{genome} or \emph{genotype}, and the actions performed by that organism are the organism's \emph{behavior}. The performance or quality of a solution is called its \emph{fitness}, and the equation, simulation, etc. that returns that fitness value is the \emph{fitness function}. The way of stochastically producing new solutions is to take an existing solution and \emph{mutate} its genome, meaning to change the genome in some random way, and or to produce a new solution descriptor by sampling portions of two parent descriptors, a process called \emph{crossover}. Solutions that produce new \emph{offspring} organisms are those that are \emph{selected}, and such selection is typically biased towards solutions with higher fitness\cite{floreano2008bio}. 

To encourage a broad exploration of the search space, many modern evolutionary algorithms encourage diversity through a variety of different techniques, including increasing mutation rates when the rate of performance improvement stagnates\cite{clune2005investigations,eiben1999parameter, clune2008natural,floreano2008bio, russell1995artificial}, explicitly selecting for genetic diversity\cite{floreano2008bio, stanley2002evolving} or behavioral diversity\cite{lehman2011abandoning, lehman2011evolving, lehman2011novelty, lehman2008exploiting, Mouret2012}, or changing the structure of the population\cite{whitley1999island}.\todoOfficialVersion{People probably won't know what we mean by the previous clause} Such diversity-promoting techniques often improve the quality of the solutions produced and the number of different types of solutions explored, but search algorithms still tend to converge to one or a few good solutions early and cease to make further progress\cite{floreano2008bio, lehman2011abandoning, Mouret2012}.

\todoOfficialVersion{Decide if we want to specifically name/list/cite some of these older techniques. See email thread ``Restricted Tournament Selection''; [JBM] it does not hurt to cite a few papers on fitness sharing.}

An alternate idea proposed in recent years is to abandon the goal of improving performance altogether, and instead select only for diversity in the feature space (also called the behavior space): This algorithm, called Novelty Search, can perform better than performance-driven search on deceptive problems\cite{lehman2011abandoning, lehman2011evolving, lehman2011novelty, lehman2008exploiting}. The user defines how to measure the distance between behaviors, and then Novelty Search seeks to produce as many different behaviors as possible according to this distance metric. The algorithm stops when an individual in the population solves the objective (i.e. their performance is high enough). Because Novelty Search does not work well with very large feature/behavioral spaces\cite{lehman2010revising,cuccu2011novelty}, there have been many proposals for combining selection for novelty and performance\cite{gomes2015devising, mouret2011novelty, cuccu2011novelty, inden2013examination}. The main focus of these hybrid algorithms remains finding the single best individual that solves a task, or a set of individuals that represent the best possible tradeoff between competing objectives. 

In the last few years, a few algorithms have been designed whose goal is not to return one individual that performs well on one objective, but a repertoire of individuals that each performs well on a different, related objective\cite{clune2013originModularity, lehman2011evolving, cully2015evolving}. Along with research into behavioral diversity and Novelty Search, such repertoire-gathering algorithms inspire the algorithm we present in this paper. 

While the exploration of search spaces is at the center of many discussions in optimization, we rarely \emph{see} these search spaces because they are often too high-dimensional to be visualized. While the computer science literature offers plenty of options for dimensionality reduction and visualization of high-dimensional data \cite{tenenbaum2000global,Haykin1998,kohonen2001self,andrews1972plots}, such algorithms are ``passive'' in that they take a fixed data set and search for the best low-dimensional visualization of it. They do not tackle the issue of generating this data set. In other words, they do not explore a high-dimensional space in such a way as to reveal interesting properties about it to a user via a low-dimensional visualization. Such exploration algorithms are necessary when the entire search space is too large to be simply visualized by a dimensionality reduction algorithm, but instead must be actively explored to learn interesting facts about it. For example, to identify all the performance peaks in a large search space, we must actively search for them. It is not enough to sample millions of solutions and plot them, for the same reason as random sampling is often not a good optimization algorithm: finding a fitness peak by chance is very unlikely for  large search spaces (in most cases, the probability of finding the best possible fitness will decrease exponentially when the number of dimensions of the search space increases).

Here we present a new algorithm that, given $N$ dimensions of variation of interest chosen by the user, searches for the highest-performing solution at each point in the space defined by those dimensions~(Fig.~\ref{conceptFig}). These dimensions are discretized, with the granularity a function of available computational resources. Note that the search space can be high-dimensional, or even of infinite dimensions, but the feature space is low-dimensional by design. We call this algorithm the multi-dimensional archive of phenotypic elites, or MAP-Elites. It was used and briefly described in\cite{cully2015robots}, but this paper is the first to describe and explore its properties in detail.

The benefits of MAP-Elites include the following:

\begin{itemize}
\item Allowing users to create diversity in the dimensions of variation they choose.

\item Illuminating the fitness potential of the entire feature space, not just the high-performing areas, revealing relationships between dimensions of interest and performance.

\item Improved optimization performance; the algorithm often finds a better solution than the current state-of-the-art search algorithms in complex search spaces because it explores more of the feature space, which helps it avoid local optima and thus find different, and often better, fitness peaks. 

\item The search for a solution in any single cell is aided by the simultaneous search for solutions in other cells. This parallel search is beneficial because (1) it may be more likely to generate a solution for one cell by mutating a solution to a more distant cell, a phenomenon called ``goal switching'' in a new paper that uses MAP-Elites\cite{nguyen2015introducing}, or (2) if it is more likely to produce a solution to a cell by crossing over two solutions from other cells. If either reason is true, MAP-Elites should outperform a separate search conducted for each cell. There is evidence that supports this claim below, and this specific experiment was conducted in Nguyen et al. 2015\cite{nguyen2015introducing}, which found that MAP-Elites does produce higher-performing solutions in each cell than separately searching for a high-performing solution in each of those cells. 

\item Returning a large set of diverse, high-performing individuals embedded in a map that describes where they are located in the feature space, which can be used to create new types of algorithms  or improve the performance of existing algorithms\cite{cully2015robots}.

\end{itemize}

\section{Optimization vs. Illumination Algorithms}
\label{illumination}

Optimization algorithms try to find the highest-performing solution in a search space. Sometimes they are designed to return a set of high-performing solutions, where members in the set are also good on other objectives, and where the set represents the solution on the Pareto front of tradeoffs between performance and quality with respect to those other objectives. Optimization algorithms are not traditionally designed to report the highest-performing solution possible in an area of the feature space that cannot produce either the highest-performing solution overall, or a solution on the Pareto front. 

A different kind of algorithm, which we call \emph {illumination algorithms}, are designed to return the highest-performing solution at each point in the feature space. They thus illuminate the fitness potential of each region of the feature space. In biological terms, they illuminate the phenotype-fitness map\cite{bull2011phenotype}. Any illumination algorithm can also be used as an optimization algorithm, making illumination algorithms a superset of optimization algorithms. MAP-Elites is an illumination algorithm. It is inspired by two previous illumination algorithms, Novelty Search + Local Competition (NS+LC)\cite{lehman2011evolving} and the Multi-Objective Landscape Exploration algorithm
(MOLE)\cite{clune2013originModularity}. All three are described below.

\begin{figure*}
\begin{algorithmic}
\Procedure{MAP-Elites Algorithm (simple, default version)}{}
\State $(\mathcal{P} \leftarrow \emptyset, \mathcal{X} \leftarrow \emptyset)$\Comment{\emph{Create an empty, $N$-dimensional map of elites:  \{solutions $\mathcal{X}$ and their performances $\mathcal{P}$\} }}
\For{iter $  = 1\to I$} \Comment{\emph{Repeat for $I$ iterations.}}
\If{iter $< G$} \Comment{\emph{Initialize by generating $G$ random solutions}}
  \State $\mathbf{x'}\leftarrow $ random\_solution()  
\Else \Comment{\emph{All subsequent solutions are generated from elites in the map}}
  \State $\mathbf{x}\leftarrow $ random\_selection($\mathcal{X}$) \Comment{\emph{Randomly select an elite $x$ from the map $\mathcal{X}$}}
  \State $\mathbf{x'}\leftarrow $ random\_variation($\mathbf{x}$) \Comment{\emph{Create $x'$, a randomly modified copy of $x$ (via mutation and/or crossover)} }
\EndIf
\State $\mathbf{b'}\leftarrow $feature\_descriptor($\mathbf{x'}$) \Comment{\emph{Simulate the candidate solution $x'$ and record its feature descriptor $\mathbf{b'}$}}
\State $p'\leftarrow $performance($\mathbf{x'}$) \Comment{\emph{Record the performance $p'$ of $x'$}}
\If{$\mathcal{P}(\mathbf{b'})= \emptyset$ or $\mathcal{P}(\mathbf{b'})<p'$}\Comment{\emph{If the appropriate cell is empty or its occupants's performance is $\leq p'$, then}}
\State $\mathcal{P}(\mathbf{b'})\leftarrow p'$ \Comment{\emph{store the performance of $x'$ in the map of elites according to its feature descriptor $\mathbf{b'}$}}
\State $\mathcal{X}(\mathbf{b'})\leftarrow \mathbf{x'}$ \Comment{\emph{store the solution $x'$ in the map of elites according to its feature descriptor $\mathbf{b'}$}}
\EndIf
\EndFor
\State \Return feature-performance map ($\mathcal{P}$ and $\mathcal{X}$)
\EndProcedure
\end{algorithmic}
\caption{A pseudocode description of the simple, default version of MAP-Elites.}
\label{mapElitesPseudocodeSimpleVersion}
\end{figure*}

\section{Details of the MAP-Elites algorithm}

MAP-Elites is quite simple, both conceptually and to implement. Pseudocode of the algorithm is in Fig.~\ref{mapElitesPseudocodeSimpleVersion}. First, a user chooses a performance measure $f(x)$ that evaluates a solution $x$. For example, if searching for robot morphologies, the performance measure could be how fast the robot is. Second, the user chooses $N$ dimensions of variation of interest that define a \emph{feature space} of interest to the user. For robot morphologies, one dimension of interest could be how tall the robot is, another could be its weight, a third could be its energy consumption per meter moved, etc. An alternate example could be searching for chess programs, where the performance measure is the win percentage, and the dimensions of variation could be the aggressiveness of play, the speed with which moves are selected, etc. A further example is evolving drug molecules, where performance could be a drug's efficacy and dimensions of variation could be the size of molecules, the cost to produce them, their perishability, etc. 

\todoOfficialVersion{Maybe somewhere in the paper, we should advise more specifically how to choose the dimensions.. E.g.

1. Choose dimensions as how you want to explore the search space.

2. Choose dimensions as to create best gradients that lead to desired solutions.}
Each dimension of variation is discretized based on user preference or available computational resources. This granularity could be manually specified or automatically tuned to the available resources, including starting with a coarse discretization and then increasing the granularity as time and computation allow. 

Given a particular discretization, MAP-Elites will search for the highest performing solution for each cell in the $N$-dimensional feature space. For example, MAP-Elites will search for the fastest robot that is tall, heavy, and efficient; the fastest robot that is tall, heavy, and inefficient, the fastest robot that is tall, light, and efficient, etc. 

The search is conducted in the \emph{search space}, which is the space of all possible values of $x$, where $x$ is a description of a candidate solution. In our example, the search space contains all possible descriptions of robot morphologies (note that we must search in the space of \emph{descriptions} of robot morphologies; it is not possible to search directly in the space of robot morphologies or directly in the feature space). We call the $x$ descriptor a \emph{genome} or \emph{genotype} and the robot morphology the \emph{phenotype}, or $p_{x}$. We have already mentioned that a function $f(x)$, called a \emph{fitness function}, returns the performance of each $x$. A feature (a.k.a. behavior) function $b(x)$ must also exist or be defined that, for each $x$, determines $x$'s value in each of the $N$ feature dimensions. In other words, $b(x)$ returns $\mathbf{b_{x}}$, which is an $N$-dimensional vector describing $x$'s features. In our example, the first dimension of $\mathbf{b_{x}}$ is the robot's height, the second dimension is its weight, and the third is its energy consumption per meter moved, etc. Some elements of the feature vector may be directly measured in the phenotype (e.g. height, weight), but others (e.g. energy consumption) require measuring the behavior of the phenotype while it performs, either in simulation or reality.
  
Note that there may be many levels of indirection between $x$ and $\mathbf{b_{x}}$. With \emph{direct encoding}, each element in the genome specifies an independent component of the phenotype\cite{stanley2003taxonomy, clune2011performance, floreano2008bio}. In that case, it is straightforward to map genotypes into phenotypes, and then measure performance and features (evaluating the phenotype in a simulator or the real world if necessary).  An extra level of indirection can occur with \emph{indirect encoding}, also known as \emph{generative} or \emph{developmental} encoding, in which information in the genome can be reused to affect many parts of the phenotype (also called \emph{pleiotropy}); such encodings have been shown to improve regularity, performance, and evolvability\cite{stanley2003taxonomy, clune2011performance, floreano2008bio,cheney2013unshackling, hornby2003generative, clune2009evolving, lee2013evolving, yosinski2011gaits, hornby2002creating, hornby2004functional}. In other words, a complex process can exist that maps genome $x \rightarrow$ to phenotype $p_{x}  \rightarrow$ to features  $\mathbf{b_{x}}$ and performance $f{x}$. 

\todoOfficialVersion{this paragraph may be irrelevant if we are not talking to the EC community.}

MAP-Elites starts by randomly generating $G$ genomes and determining the performance and features of each. In a random order, those genomes are placed into the cells to which they belong in the feature space (if multiple genomes map to the same cell, the highest-performing one per cell is retained). At that point the algorithm is initialized, and the following steps are repeated until a termination criterion is reached. (1) A cell in the map is randomly chosen and the genome in that cell produces an offspring via mutation and/or crossover. (2) The features and performance of that offspring are determined, and the offspring is placed in the cell if the cell is empty or if the offspring is higher-performing than the current occupant of the cell, in which case that occupant is discarded. 

\todoOfficialVersion{Joost writes: One thing I have always wondered, why do you call it an archive, and not a map? I mean, it is MAP-Elites, not Archive-Elites, and the archive does not behave like an archive anyway (in contrast to novelty search, where the term archive makes a lot more sense).}
The termination criterion can be many things, such as if a set amount of time expires, a fixed amount of computational resources are consumed, or some property of the archive is produced. Examples of the latter could include a certain percentage of the map cells being filled, average fitness in the map reaching a specific level, or $n$ solutions to a problem being discovered. 

One can consider the archive, which is the set of descriptors in all the cells, as the traditional \emph{population} in an evolutionary algorithm. The difference is that in MAP-Elites each member of the population is by definition diverse, at least according to the dimensions of the feature space. 

The above description, for which pseudocode is provided in Fig.~\ref{mapElitesPseudocodeSimpleVersion}, is the default way to implement MAP-Elites. To encourage a more uniform exploration of the space at a coarse resolution, and then a more fine-grained search afterwards, we created a hierarchical version that starts with larger cells in the feature space that are then subdivided into smaller cells during search after predetermined numbers of evaluations have been performed (Methods). \todoOfficialVersion{Report whether preliminary experience show this is a good idea, and according to which criteria} We further parallelized this algorithm to run on clusters of networked computers, by farming out batches of evaluations to slave nodes, instead of performing each evaluation serially (Methods). Section \ref{alternateVersions} contains ideas for additional, alternate possible variants of MAP-Elites. 

There are two things to note about MAP-Elites:
\begin{itemize}
\item{It is not guaranteed that all cells in the feature space will be filled, for two reasons. (1) There may be no genome $x$ that maps to a particular cell in the feature space. For example, it may be impossible to have a robot of a certain height and weight due to physical laws. (2) The search algorithm may fail to produce a genome that maps to a cell in the feature space, even if such a genome exists.}

\item{There are many genotypes that can be mapped to the same cell in the feature space, perhaps an infinite number. For example, there are many robot blueprints that produce a robot with the same height, weight, and energy consumption. For that reason, and because it is not known a priori which genomes will map to which cells, it is not possible to search directly in the feature space. Recall that there is, even with direct encodings, and especially with indirect encodings, a complex mapping from genome $x$ to the feature vector $\mathbf{b_{x}}$. If it is possible in a given problem to directly take steps in the feature space, then MAP-Elites is unnecessary because one could simply perform exhaustive search in the feature space. One can think of MAP-Elites as a way of trying to perform such an exhaustive search in the feature space, but with the additional challenge of trying to find the highest-performing solution for each cell in that feature space.}
\todoOfficialVersion{Also very EC-oriented.}

\end{itemize}

\section{Differences between MAP-Elites and previous, related algorithms}

In 2011, Lehman and Stanley\cite{lehman2011evolving} note that combining a selective pressure for feature diversity with one performance objective that all of the different types of phenotypes compete on is akin to having butterflies and bacteria compete with bears and badgers on one performance criterion (e.g. speed). Doing so is unhelpful for producing the fastest of each type of creature, given the different speed scales these creatures exhibit. Instead, Lehman and Stanley propose encouraging diversity in the feature space, but having each organism compete on performance only with other organisms that are near it in the feature space, an algorithm they call \emph{Novelty Search + Local Competition} (NS+LC)\cite{lehman2011evolving}. 
NS+LC accomplishes these goals via a multi-objective algorithm with two objectives: (1) maximizing an organism's performance relative to its closest 15 neighbors (i.e. local competition, but note that these relative scores are then entered into a global competition, the implications of which are discussed below), and (2) maximizing a novelty objective, which rewards organisms the further they are in feature space from their 15 closest neighbors. 
Whereas normally evolutionary algorithms do not produce much diversity within one run, but instead have to perform multiple, independent runs to showcase diversity\cite{cheney2013unshackling}, NS+LC produces a substantial amount of different types of high performing creatures within one evolving population\cite{lehman2011evolving}. 

NS+LC inspired us to create two algorithms that also seek to find the highest performing solution at each point in a feature space. The first was the Multi-Objective Landscape Exploration (MOLE) algorithm\cite{clune2013originModularity} and the second is MAP-Elites, the algorithm presented in this paper. MOLE has two objectives: the first is performance, and the second for each organism to be as far from other organisms as possible, where distance is measured in a feature space that a user specifies. 

Both NS+LC and MOLE have similar goals to MAP-Elites: they search for the highest-performing solution at each point in a feature space. However, both are more complicated and, as will be shown in the results section, do not perform as well as MAP-Elites empirically.

Specific differences between MAP-Elites and NS+LC include:
\begin{itemize}

\item Novelty Search needs to compute the feature distance to every other organism each generation; such nearest neighbor calculations are $O(n \operatorname{log}(n))$\cite{friedman1977algorithm} each generation. MAP-Elites only needs to look up the current occupant of the cell, which is $O(1)$.

\item Novelty Search contains both a current population and an archive of previous solutions that serves as a memory of which points in the feature space have been visited. Maintaining both a population and an archive requires many additional parameters that have to be chosen carefully or performance can be detrimentally affected\cite{gomes2015devising}.

\item Given that Novelty Search rewards individuals that are distant from each other in the feature space, having only a population would lead to ``cycling'', a phenomenon where the population moves from one area of the feature space to a new area and back again, without any memory of where it has already explored. The archive in NS+LC limits, but does not eliminate, this phenomenon. MAP-Elites does away with the archive vs. population distinction by having only an archive. It thus avoids cycling and is always simultaneously focused on expanding into new niches (until there are none left) and improving the performance of existing niches. 

It is thus quite easy to intuit what the selection pressure for MAP-Elites is over time. In contrast, the selection pressures for Novelty Search are more dynamic and thus harder to understand, even for Novelty Search variants that have only an archive and no population\cite{gomes2015devising}. For example, it is hard to predict how much search will be focused in each area of the feature space, because a relatively sparse area during one era of the search can become relatively crowded later on, and vice versa. 

The dynamics of NS+LC are even more dynamic, complex, and unpredictable. One thing to keep in mind is that, while the performance of solutions in NS+LC is only judged versus neighbors, these relative performance scores are then competed globally within the (relative) performance objective. Overall, therefore, NS+LC biases search towards under-explored areas of the feature space (taking into account the archive and the population), areas of the search space with the highest relative performance, and tradeoffs between these two objectives. An organism in an area that is better than its neighbors, but where this gap is not as large as elsewhere, will not be explored as often unless or until that larger performance gap elsewhere is reduced. The focus of the (relative) performance objective is thus complex and ever-changing. The diversity objective is also complex and dynamic, because NS+LC does not only store one solution per cell. Many solutions can pile up in one area of the space, creating a pressure to explore under-explored areas until those areas are more explored relative to the initial area, creating a pressure to return to the initial area, and so on. 

For both objectives, thus, it is hard to intuit both the dynamics themselves and what effects these dynamics have on search. MAP-Elites, in contrast, produces offspring by uniformly sampling from the existing archive, such that the only thing that changes over time is the number of cells that are filled and their performance. MAP-Elites thus embodies the main principle of illumination algorithms, which is to search for the highest-performing solution at each point of the feature space, in a more simple, intuitive, and predictable way. 

\item In the default version of MAP-Elites, organisms only compete with the organism (the current occupant) in their cell, so the range of features they compete with is fixed. In Novelty Search and NS+LC, organisms compete with their nearest neighbors in the feature space. Especially at the beginning of the run before the archive fills up, that might mean that organisms are competing with others that have very different features, which is contrary to the spirit of local competition in the feature space.

\end{itemize}  

Specific differences between MAP-Elites and MOLE include:
\begin{itemize}
\item MOLE features one global performance competition (via the performance objective). Thus, a few high-performing individuals will dominate this objective, making it hard to recognize and keep a slightly better performing solution in a low- or medium-performing region of the space. MAP-Elites is better at recognizing and keeping any improvement to fitness in any region of the space, no matter how the performance of that cell compares to other cells. As an example of when MOLE might fail to reward an important innovation, imagine a new solution in a medium-performing, densely packed region of the space, that is higher-performing than anything previously found in that cell. This new solution, which represents the best performance yet found in that cell, would not be selected for because it is neither high-performing versus other organisms in the population, nor would it be kept because it is diverse. Thus, the organism does not perform well in either of the MOLE objectives, yet it is precisely what we truly want: the highest performing individual found so far in that area of the feature space. 

\item Like Novelty Search, the diversity objective in MOLE has unstable temporal dynamics. The population may rush to a relatively unexplored area, fill it up, then rush off to a new relatively unexplored area, and then rush back to the original area. It does not evenly search for improvements to all areas of the map simultaneously. 

\todoOfficialVersion{add any more? JBM: care to add any?}
\end{itemize} 

\todoOfficialVersion{JBM: Do we need one of these lists for MAP-Elites vs. performance+behavioralDiversity-MOO? How about vs. plain NS?}

\section{Criteria for Measuring the Algorithms}
\label{criteria}
There are many different ways to quantify the quality of illumination algorithms and optimization algorithms. In this paper, we evaluate algorithms on the following quantifiable measures: 

\begin{itemize}

\item{\textbf{Global Performance:} For each run, the single highest-performing solution found by that algorithm anywhere in the search space divided by the highest performance possible in that domain. If it is not known what the maximum theoretical performance is, as is the case for all of our domains, it can be estimated by dividing by the highest performance found by any algorithm in any run. This measure is the traditional, most common way to evaluate optimization algorithms. One can also measure whether any illumination algorithm also performs well on this measurement. Both the ideal optimization algorithm and the ideal illumination algorithm are expected to perform perfectly on this measure.}

\item{\textbf{Global reliability:} For each run, the average across all cells of the highest-performing solution the algorithm found for each cell (0 if it did not produce a solution in that cell) divided by the best known performance for that cell as found by any run of any algorithm. Cells for which no solution was found by any run of any algorithm are not included in the calculation (to avoid dividing by zero, and because it may not be possible to fill such cells and algorithms thus should not be penalized for not doing so). Section \ref{methodsGlobalReliability} provides the formal equation. 

This measure assesses how reliable an algorithm is at finding the highest-performing solution for each cell in the map. It is the most important measure we want an illumination algorithm to perform well on, and the ideal illumination algorithm would perform perfectly on it. There is no reason to expect pure optimization algorithms, even ideal ones, to perform well on this criterion.
}

\item{\textbf{Precision (opt-in reliability):} For each run, if (and only if) a run creates a solution in a cell, the average across all such cells of the highest performing solution produced for that cell divided by the highest performing solution any algorithm found for that cell. Section \ref{methodsPrecision} provides the formal equation.

This metric measures a different notion of reliability, which is the trust we can have that, if an algorithm returns a solution in a cell, that solution will be high-performing relative to what is possible for that cell. To anthropomorphize, the algorithm gets to opt-in which cells it wishes to fill and thus be measured on. The ideal illumination algorithm would have a perfect score of 1 for this criterion. Optimization algorithms should fare better on this criterion than global reliability, because they will tend to explore only a few areas of the feature space, but should produce high-performing solutions in many cells in those areas. Note, however, that if an optimization algorithm starts in a low-performing region of the feature space and moves to a neighboring region, it is expected that its relative performance in the cells it started in will stay low, as optimization algorithms are not asked to improve performance in those cells. Thus, even ideal optimization algorithms are not expected to perform perfectly on this criterion on average, although they may do so once in a while.
}

\item{\textbf{Coverage:} Measures how many cells of the feature space a run of an algorithm is able to fill of the total number that are possible to fill. The mathematical details are specified in section \ref{methodsCoverage}. This measure does not include the performance of the solutions in the filled cells. The ideal illumination algorithm would score perfectly on this metric. The ideal optimization algorithm is not expected to perform well on this criterion.}

\end{itemize}

\section{Experiments and Results}
We evaluated MAP-Elites in three different search spaces: neural networks, simulated soft robot morphologies, and a real, soft robotic arm. The neural network search space is interesting because evaluations are fast, allowing us to draw high-resolution feature-space maps for a high-dimensional search space. The experiments with both simulated and real soft robot are interesting because soft robots are important, new design spaces where traditional design and control methods do not work well, if at all. Thus, we need advanced search algorithms to find high-performing designs. The first two search spaces (neural networks and simulated soft robots) are extremely high-dimensional, demonstrating the ability of MAP-Elites to create low-dimensional feature maps from high-dimensional search spaces. The third, involving the soft robot arm,  involves evaluations that are performed directly on a real robot because the soft properties of the robot are too complex to simulate. This domain demonstrates that MAP-Elites is also effective even in a low-dimensional, challenging, real-world problem. 

\subsection{Search space 1: neural networks}

\todoOfficialVersion{add background on how the space of neural networks is super high D, so we need stochastic search algorithms to find their weights (and possibly topology) when doing RL, as opposed to supervised learning, where backprop can work}

This problem domain is identical to one from Clune et al. 2013\cite{clune2013originModularity}, which itself is based on the domain from Kashtan and Alon 2005\cite{kashtan2005spontaneous}. The following explanation of the domain is adapted from Clune et al. 2013\cite{clune2013originModularity}.
  
The problem involves a neural network that receives stimuli
from an eight-pixel retina. Patterns shown on the retina's left and right
halves may each contain an object (i.e. a pattern of interest). Networks have to answer
whether an object is present on both the left \emph{and} right sides
of the retina\cite{kashtan2005spontaneous, clune2013originModularity}. Each network iteratively sees all possible 256 input patterns and answers true ($\geq0$) or false ($<0$). Its performance is the percentage of correct answers. 

\todoOfficialVersion{in the following paragraph you say squared length...confirm with JBM that it is squared}
Because it has been shown that minimizing connection costs promotes the evolution of modularity\cite{clune2013originModularity}, it is interesting to visualize the relationship between network connection costs and modularity. To do so, we can create a 2D feature space where the first feature dimension ($x$ axis) is connection cost (the sum of the squared length of the connections in a network\cite{clune2013originModularity}), and the second feature dimension is network modularity (computed using an efficient approximation of Newman's modularity score\cite{Leicht2008}). The resolution of the map is 512 $\times$ 512; the map is filled by the hierarchical version of MAP-Elites with 10,000 evaluations (Methods). 

For this domain, we compare MAP-Elites to three other algorithms: (1) a traditional, single-objective evolutionary algorithm with selection for performance only, which thus does not explicitly seek diversity in either feature dimension, (2) novelty search with local competition (NS+LC) \cite{lehman2011evolving}, which is described above, where novelty is measured in the same 2D feature space, and (3) to random sampling. \todoOfficialVersion{shouldn't  we compare to MOLE too; seems like the best/only place for us to show why MAP-Elites is better than MOLE}For these three control experiments, we record all the candidate solutions evaluated by the algorithm and then keep the best one found per cell in the feature space (i.e. the \emph{elite} performer for each cell), and report and plot these data. Each treatment is allocated the same number of fitness evaluations (Methods). For each treatment, 20 independent runs are performed, meaning 20 independent replicates that each start with a different random number seed and thus have different stochastic events. 

\begin{figure*}[ht!]
     \begin{center}
        \subfigure{
            \label{fig:first}
            \includegraphics[width=\textwidth]{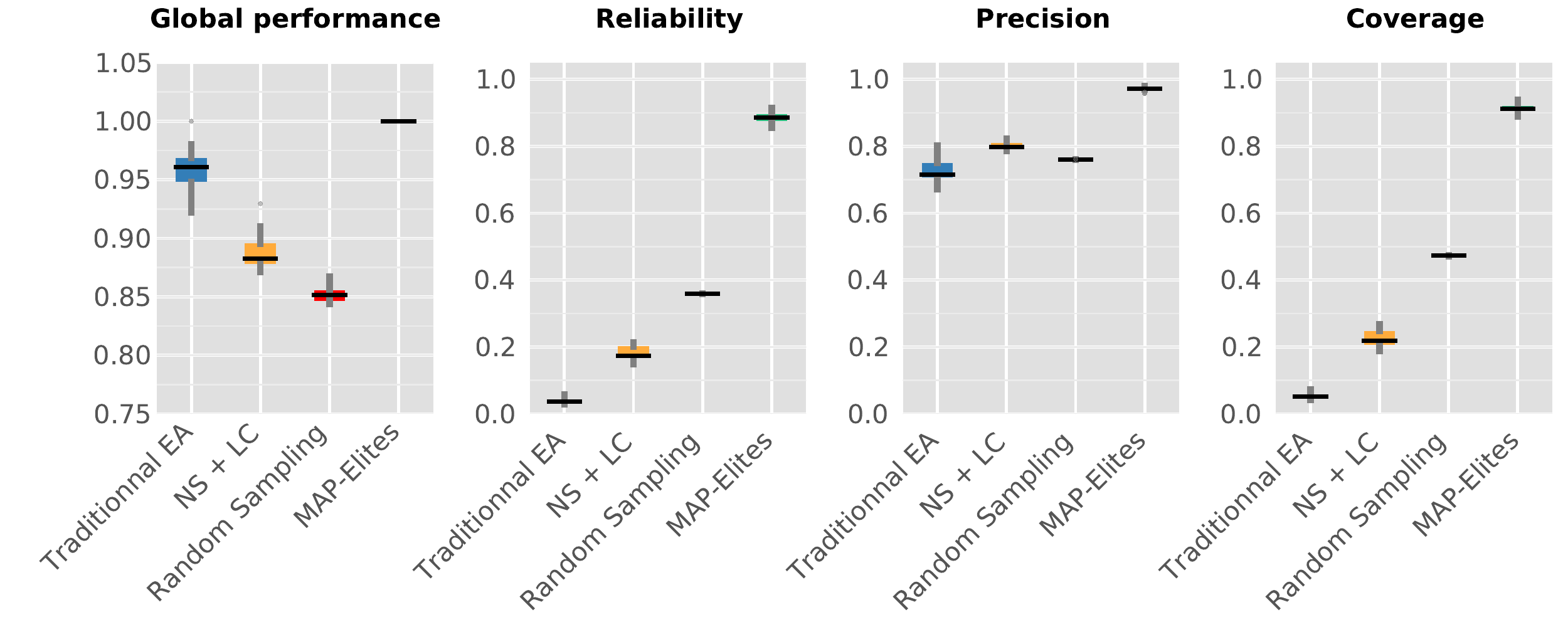}
        }\\
        \subfigure[Traditional EA]{
           \label{fig:second}
           \includegraphics[width=0.3\textwidth]{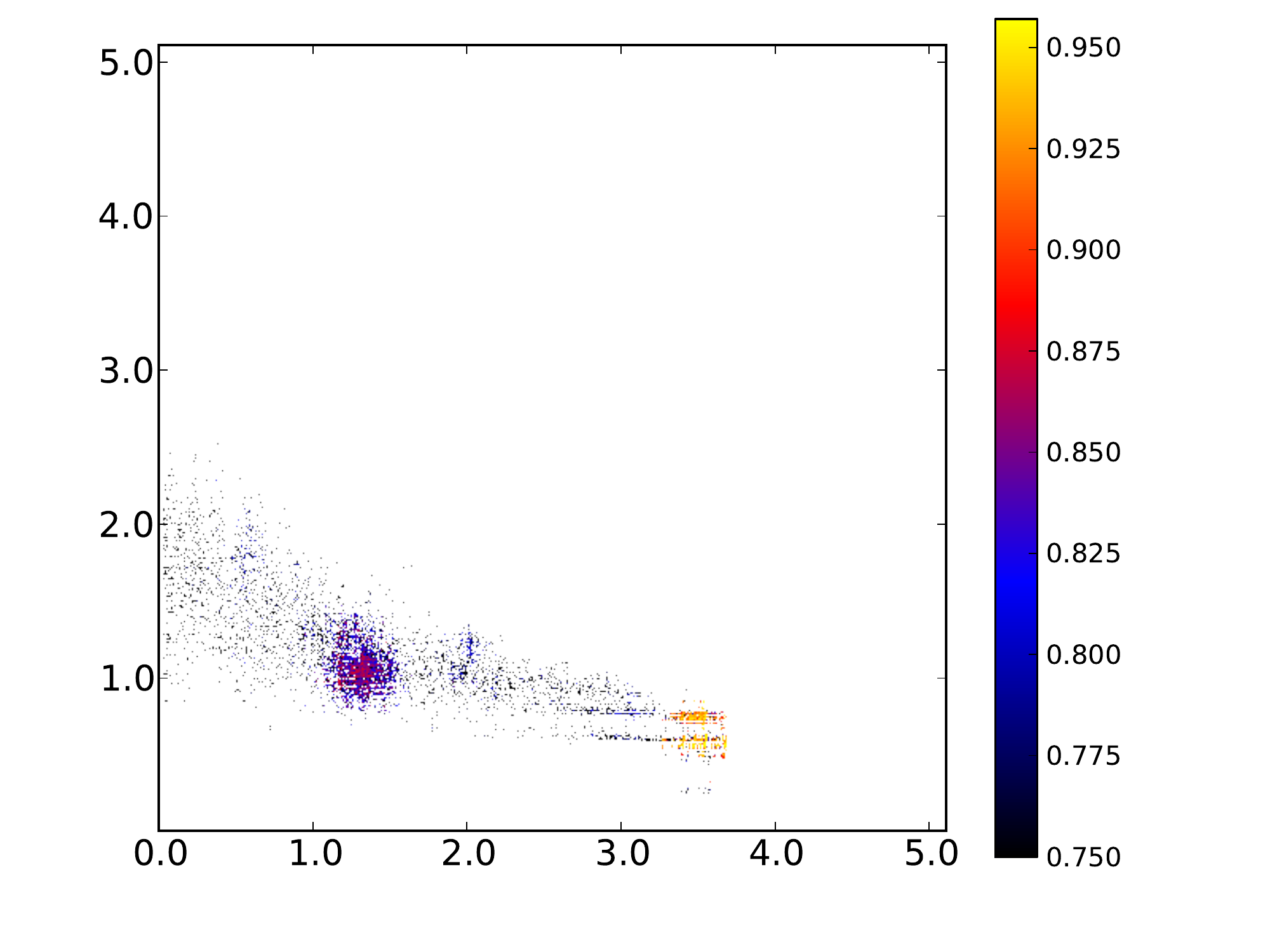}
        } 
        \subfigure[Novelty Search + Local Competition]{
            \label{fig:third}
            \includegraphics[width=0.3\textwidth]{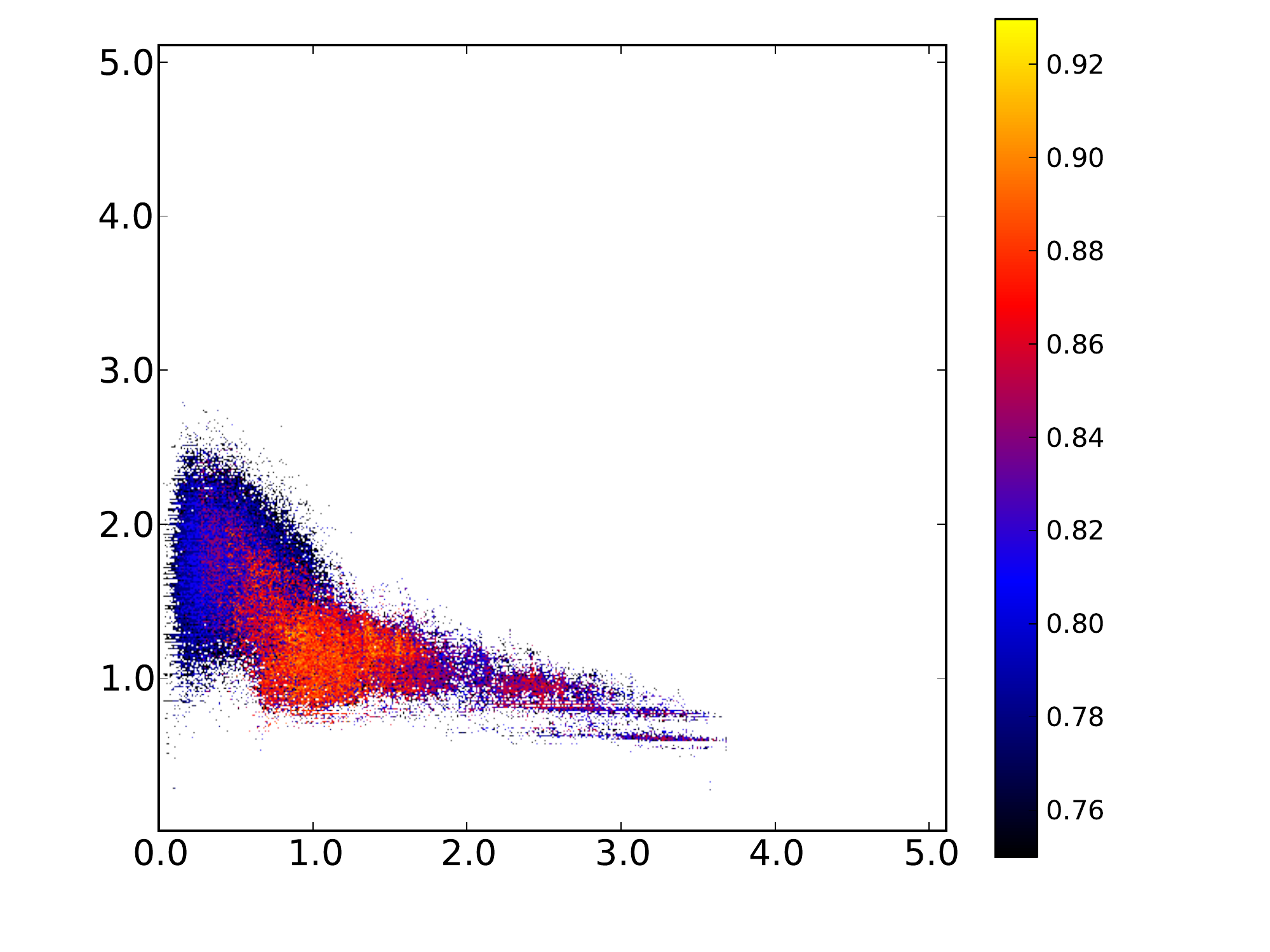}
        }
        \subfigure[Random Sampling]{
            \label{fig:third}
            \includegraphics[width=0.3\textwidth]{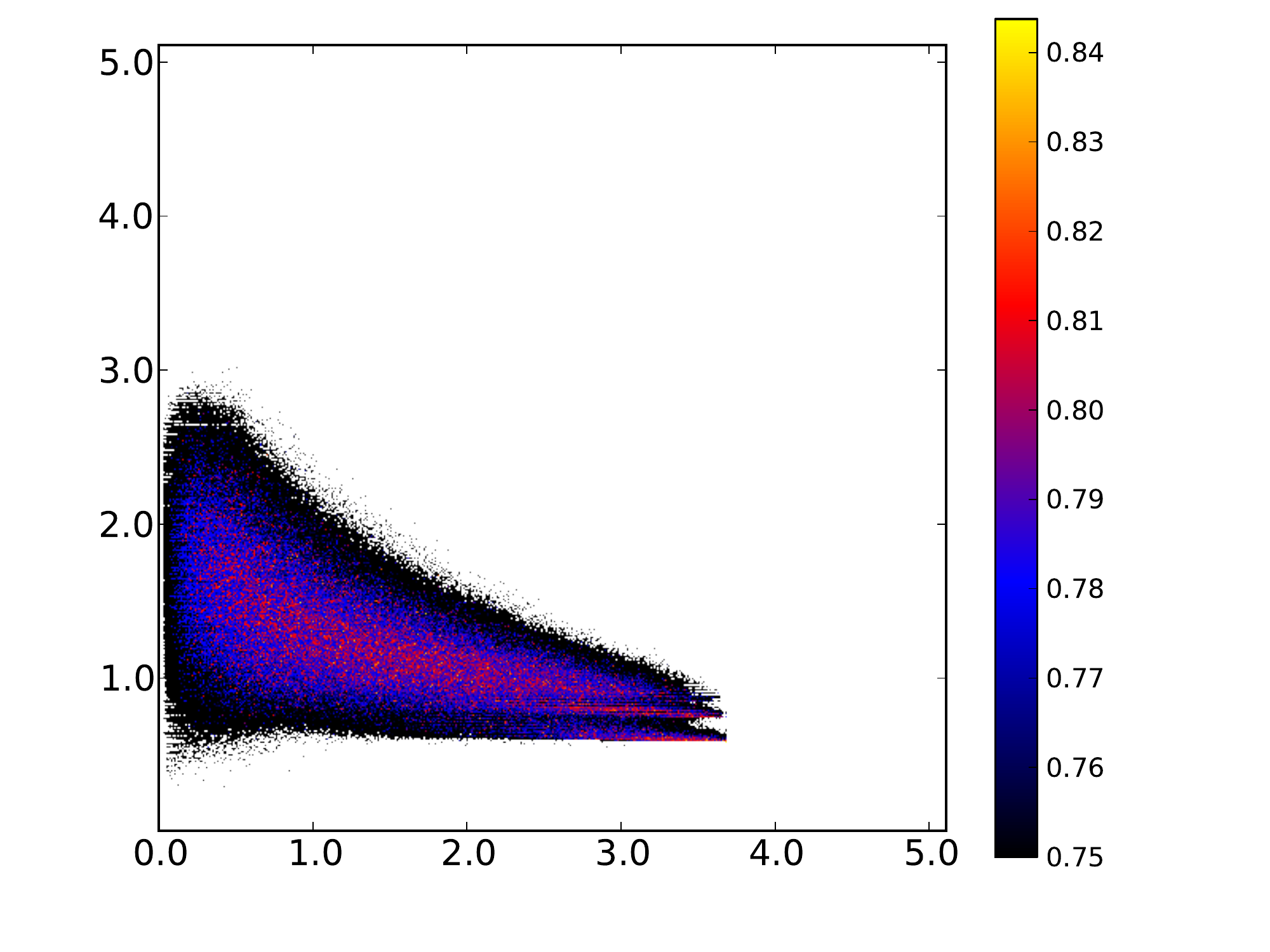}
        }
        \subfigure[MAP-Elites]{
            \label{fig:fourth}
            \includegraphics[width=0.3\textwidth]{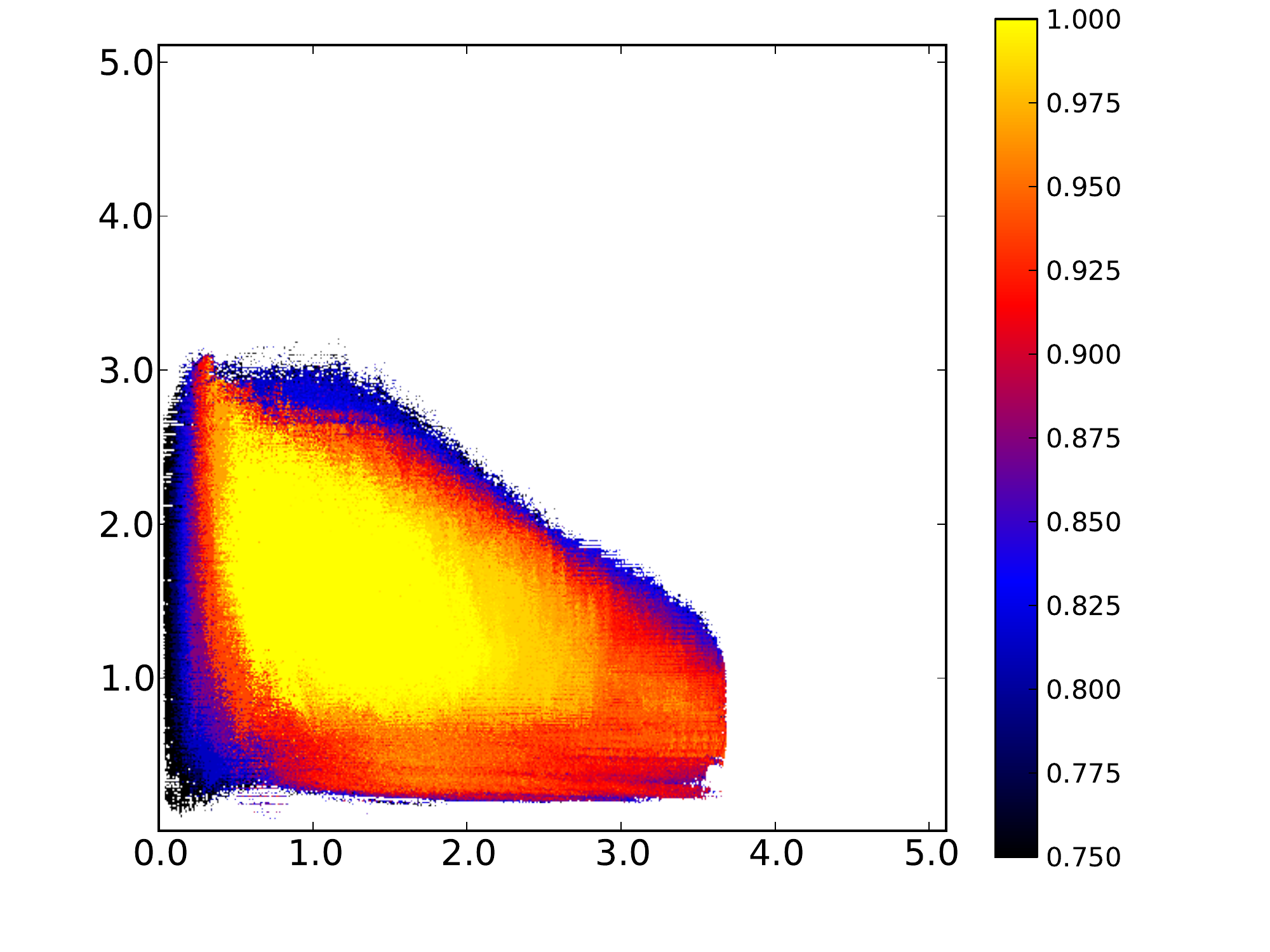}
        }
    \end{center}
    \caption{MAP-Elites produces significantly higher-performing and more diverse solutions than control algorithms. \textbf{Top:} MAP-Elites significantly outperforms controls on global performance (finding the single highest-performing solution), reliability (average performance across all fillable cells), precision (average performance only of cells filled by the algorithm), and coverage (the number of cells filled). All of the metrics are normalized. Section \ref{criteria} explains these metrics in more detail. In the box plot for each metric, the black line shows the median\todoOfficialVersion{explain what the colored box and gray box cover: 25-75 and 5-95?}. \textbf{Bottom:} Example maps produced by a single run (the first one) of each treatment. As described in Clune et al. 2013\cite{clune2013originModularity}, the $x$-axis is connection cost, the $y$-axis is modularity, and heat map colors indicate normalized performance. These maps show that MAP-Elites illuminates more of the feature space, revealing the fitness potential of each area.}
\label{fig:retina_results}

\end{figure*}

The results reveal that MAP-Elites scores significantly higher ($p < 1 \times 10^{-7}$) than the three control algorithms on all four criteria described in section \ref{criteria}: global performance, global reliability, precision, and coverage (Fig. \ref{fig:retina_results}, Top). Qualitatively, the difference in MAP-Elites vs. the controls is apparent in typical, example maps produced by each treatment (Fig. \ref{fig:retina_results}, Bottom).  Overall, MAP-Elites finds solutions that are more diverse and high-performing than traditional optimization algorithms (here called the ``traditional EA''), novelty search with local competition, and random sampling. 
\todoOfficialVersion{We should compare to NS too, and MOLE}

It is surprising that, even when looking only at the best performance overall (global performance), MAP-Elites outperforms the traditional EA, which focuses explicitly on finding the single best-performing individual in the search space. That is likely because the retina problem is deceptive\cite{clune2013originModularity} and this traditional evolutionary algorithm has no pressure for diversity, which is known to help with deception\cite{floreano2008bio}. 

While MAP-Elites significantly outperforms all controls on both reliability and precision (opt-in reliability), the gap is much narrower for precision, as is to be expected. 
In terms of coverage, random sampling was the second best of the algorithms in our study. MAP-Elites likely outperforms it in this regard because mutations to members of a diverse population are more likely to fill new cells versus randomly generating genomes. That is especially true if cells are more likely to be filled by mutating a nearby cell than by randomly sampling from the space of all possible genotypes. Imagine, for example, that most randomly sampled genotypes are in the center of a map. In that case, it would be unlikely to produce an organism in a corner by random sampling. In contrast, MAP-Elites could slowly accumulate organisms in cells closer and closer to the corner, making it more likely to eventually fill that corner. Random sampling likely produces more coverage than the traditional EA because the latter tends to allocate new individuals as offspring of the highest-performing organisms found so far, focusing search narrowly at the expense of exploring the feature space. It is not clear why random sampling produced more coverage than NS+LC, although this result needs to be tested across a wider range of parameters before its robustness is known. 
\\
\indent We can also report anecdotally that MAP-Elites performs much better in this domain than the MOLE algorithm, which was previously applied to this same domain and feature space\cite{clune2013originModularity}. For this early draft of the paper we do not yet have data to share because the MOLE runs in that paper were at a lower resolution; we will add a fair comparison of MOLE to MAP-Elites in a future draft of this paper. We can report that the MOLE figures from Clune et al. 2013 required merging data from many (specifically, 30) different runs of MOLE, meaning that across many MOLE runs we took the highest-performing network found in each cell. The variance in these MOLE runs was high, such that many of the runs did not find high-performing networks in large regions of the space; we thus were only able to get a good picture of the fitness potential of each region by taking data from many different runs. That high variance means that any individual MOLE run did not produce a reliable, consistent, true picture of the fitness potential of each region of the space; such a picture only came into view with a tremendous amount of computation spent on many MOLE runs. In contrast, each individual MAP-Elites run produces a consistent picture that looks similar to the result of merging many MOLE runs. There is still variance between MAP-Elites runs, but it is much smaller, meaning that each run of the algorithm is more reliable. 

\begin{figure*}[ht!]
     \begin{center}
        \subfigure{
            \label{fig:first}
            \includegraphics[width=0.99\columnwidth]{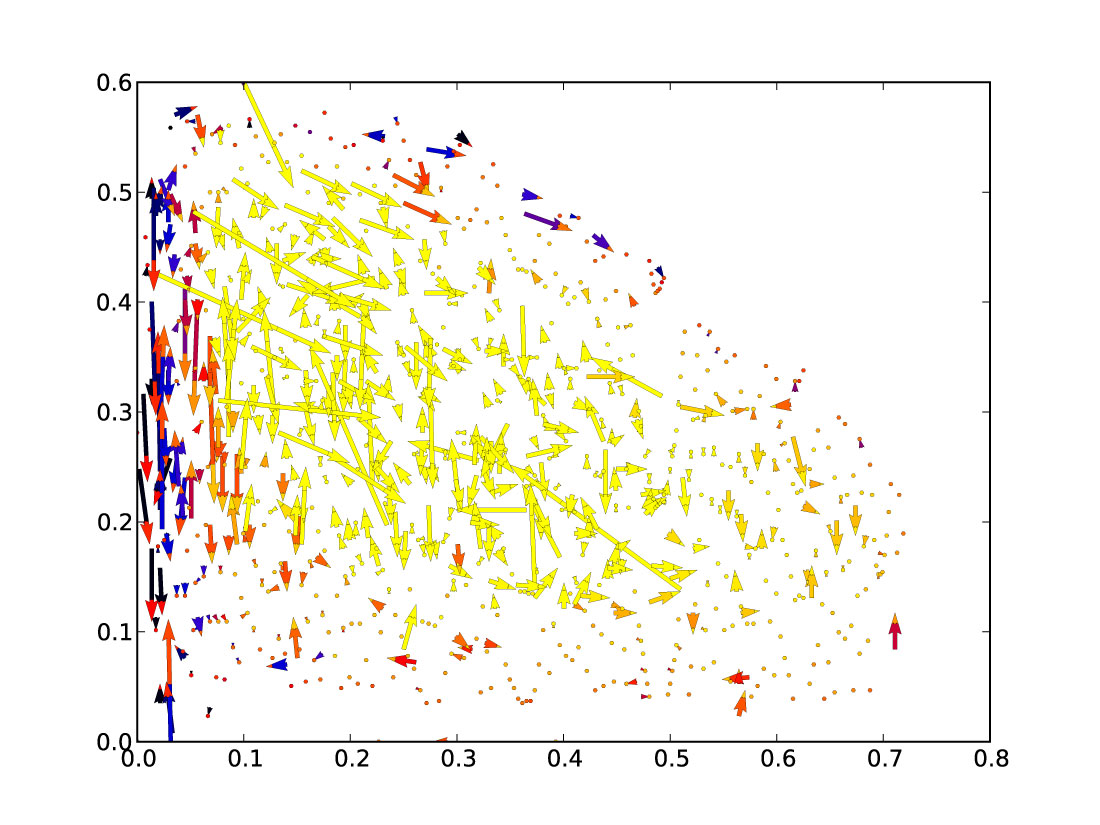}
        }
         \subfigure{
            \label{fig:first}
            \includegraphics[width=0.99\columnwidth]{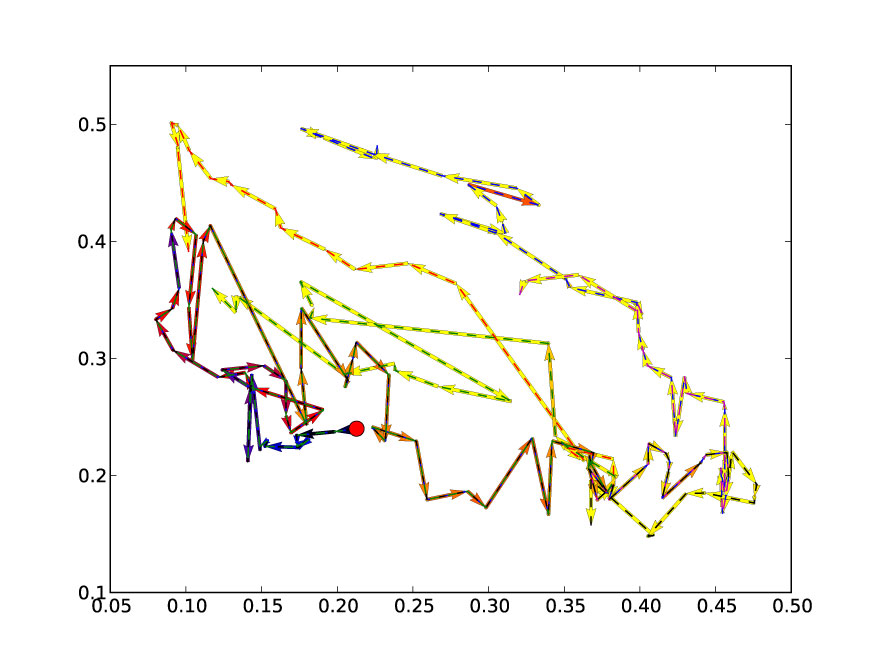}
        }
    \end{center}
    \caption{%
        Most elites are found by mutating a parent genome that was nearby in the feature space, but the entire lineages of example elites reveals search paths that traverse large distances through the feature space. The data in these plots are from the neural network domain. As in Fig.~\ref{fig:retina_results} and Fig.~3 of Clune et al.\cite{clune2013originModularity}, the $x$-axis is connection cost and the $y$-axis is modularity.  \textbf{Left:} For a random subset of elites from the neural network domain, we draw an arrow pointing at that elite that starts in the location of the parent that produced that elite. If there were no correlation between the location of an elite and its parent, there would be far more long arrows. Most elites are produced from parents within a range of distances in a nearby region (approximately 0.2 or less). The color of the beginning of each arrow denotes the performance of the parent, and the color toward the tip of the arrow denotes the performance of the elite. Note that many high-performing elites descend from other high-performing elites, but often not from direct neighboring cells. These data suggest that collecting high-performing elites in many different locations helps discover high-performing elites in new locations, which is likely why MAP-Elites is able to find so many different, high-performing solutions. \textbf{Right:} Example lineages tracing all of the descendants of four randomly selected final elites. For each of the four elites, a dashed line of a different color (green, orange, blue, or purple) starts at its randomly generated, generation 0 ancestor (red circle), which interestingly is the same for all four elites. Note that the colors and paths are harder to differentiate when the different lineages overlap. Each dashed line passes through the location in the feature space of each ancestor along the lineage of that elite and terminates at that elite's location in the feature space. 
The color of arrows along each lineage denote the performance of the parent that was located at the tail end of the arrow and produced the offspring at arrowhead. The main conclusion is that the stepping stones that lead to a high-performing elite at a particular location in the feature space are distributed throughout the feature space, suggesting that MAP-Elites' strategy of simultaneously rewarding high-performing organisms at each point in the space may help discover high-performing solutions in very different regions.
        \todoOfficialVersion{I think the right figure is unreadable. Can we try a few different things? Options: (1) Different color per path, with a circle at the beginning and an arrow at the end. (2) Different color per path, but different shades of that color (lighter to darker) to indicate the start-to-finish of the lineage. (3) One path per panel, 4 panels?}
     }%
     \todoOfficialVersion{Add heatmap}
       \label{fig:retina_paths}
\end{figure*}

We next investigated the assumption that elites are found by mutating genomes nearby in the feature space, and found that this assumption is largely true  (Fig. \ref{fig:retina_paths}, Left). Most organisms descend from nearby organisms, whether close neighbors, nearby neighbors, or more distant neighbors within the same region of the space. None of the organisms we randomly sampled were produced by a parent more than halfway across the feature map. That said, many high-performing elites do descend, not from immediate neighbors, but from a high-performing neighbor a medium distance away. That fact shows that purely local search, which likely concentrates on one area of the feature space, may not be the best way to discover high-performing solutions, and suggests that one reason MAP-Elites is able to find so many high-performing solutions is because collecting a large reservoir of diverse, high-performing solutions makes it more likely to find new, different, high-performing solutions. 

Looking at the direct parents of elites suggests that a relatively local, but overlapping, search is taking place in each region of the map. However, looking at the entire lineage of four randomly chosen elites reveals that lineages frequently traverse long paths through many different regions of the map~(Fig. \ref{fig:retina_paths}, right). These lineages further underscore the benefit of simultaneously searching for high-performing organisms at each point in the map: doing so may provide stepping stones to high-performing solutions in a region that may not have been discovered had search been trying to increase performance by searching only in that region. This result was replicated in a recent study in a different domain that investigated this issue with MAP-Elites in more depth\cite{nguyen2015introducing}.

\subsection{Simulated soft, locomoting robot morphologies}

Soft robots are made of soft, deformable materials; they open up new design spaces, allowing the creation of robots that can perform tasks that traditional robots cannot\cite{trivedi2008soft, ilievski2011,lipson2014challenges, laschi2012}. For example, they can adapt their shape to their environment, which is useful in restricted spaces like pipelines, caves, and blood arteries. They are also safer to have around humans\cite{bicchi2004}. However, they are harder to design because their components have many more non-linear degrees of freedom\cite{Hiller2012, lipson2014challenges}.

It has previously been shown that an evolutionary algorithm with a modern, generative encoding (explained below) can produce a diversity of soft robots morphologies that move in different ways\cite{cheney2013unshackling}. However, the diversity of morphologies shown in that paper and its accompanying video (\url{https://youtu.be/z9ptOeByLA4}) came from \emph{different runs} of evolution. Within each run, most of the morphologies were similar. As is typical with evolutionary algorithms, in each run the search found a local optimum and became stuck on it, spending most of the time exploring the similar designs on that peak. 

The morphologies evolved in Cheney et al.\cite{cheney2013unshackling} also rarely included one of the four materials available, a stiff (dark blue) material analogous to bone. The authors (one of which is the last author on this paper) tried many different parameters and environmental challenges to encourage the optimization algorithm to use more of this material, but it rarely did. One could, of course, explicitly include a term in the fitness function to reward the inclusion of this material, but that may cause evolution to over invest in it, and it is hard to know ahead of time how much material to encourage the inclusion of to produce interesting, functional designs. The ideal would be to see the highest-performing creature at each level of bone use, and thus learn how the use of bone affects both fitness and morphology design. That is exactly what MAP-Elites is designed for. The authors of Cheney et al. 2013 were also interested in morphologies of different sizes, which can also be added as a different dimension of variation to be explored by MAP-Elites. 

Here we test whether MAP-Elites can address the issues raised in the two previous paragraphs. Specifically, we test (1) whether MAP-Elites can produce a large diversity of morphologies within one run and (2) whether it can produce high-performing morphologies for a range of levels of bone use and body size, and combinations thereof.  

We adopt the same domain as Cheney et al. 2013\cite{cheney2013unshackling} by evolving multi-material, soft robots in the Voxcad simulator\cite{hiller2014dynamic}. Robots are specified in a space of 10 $\times$ 10 $\times$ 10 voxels, where each voxel is either empty or filled with one of four kinds of material: bone (dark blue,  stiff), soft support tissue (light blue, deformable), muscles that contract and expand in phase (green, cyclical volumetric actuation of 20\%), and muscles that contract and expand in opposite phase (red, counter-cyclical volumetric actuation of 20\%). 

The material of each voxel is encoded with a compositional pattern-producing network (CPPN)\cite{stanley2007compositional}, an encoding based on developmental biology that causes robot phenotypes to be more regular and high-performing\cite{stanley2007compositional, stanley2009hypercube, gauci2010autonomous, clune2011performance, yosinski2011gaits, lee2013evolving, cheney2013unshackling, cheney2014evolved,tarapore2015evolvability}.  
CPPNs are similar to neural networks, but with evolvable activation functions (in this paper, the functions can be sine, sigmoid, Gaussian, and linear) that allow the network to create geometric patterns in the phenotypes they encode. Because these activation functions are
regular mathematical functions, the phenotypes produced by CPPNs tend to
be regular (e.g. a Gaussian function can create symmetry and a sine function can create repetition). CPPN networks are genomes that are run iteratively for each voxel in the workspace to determine whether that voxel is empty or full and, if full, which type of material is present. Specifically, for each voxel, the Cartesian ($x$, $y$, and $z$) coordinates of the voxel and its distance from the center ($d$) are provided as inputs to the CPPN, and one CPPN output specifies whether a voxel is empty. If the voxel is not empty, the maximum value of an additional four outputs (one per material type) determines the type of material for that voxel. This method of separating the presence of a phenotypic component and its parameters into separate CPPN outputs has been shown to improve performance\cite{verbancsics2011constraining, huizinga2014evolving}.  If there are multiple disconnected voxel patches, only the most central patch is considered as the robot morphology. A lengthier explanation of CPPNs and how they specify the voxels of the soft robots in this domain can be found in Cheney et al. 2013\cite{cheney2013unshackling}, from which some text in this description of methods was derived. 

While the soft robot morphologies are indirectly encoded by CPPNs, the CPPN networks themselves are directly encoded and evolved according to the principles of the NEAT algorithm\cite{stanley2002evolving}, as is customary for CPPNs\cite{stanley2007compositional, stanley2009hypercube, gauci2010autonomous, clune2011performance, yosinski2011gaits, lee2013evolving, cheney2013unshackling, cheney2014evolved}. Here, the NEAT principles are implemented in the Sferes$_{v2}$\cite{mouret2010sferesv2} evolutionary platform, which has some departures from the original NEAT algorithm. Specifically, our direct encoding does not include crossover or genetic diversity via speciation. See Mouret and Doncieux 2012\cite{Mouret2012} for a more detailed description of the Sferes version of NEAT.

Performance for these soft robots is defined as the distance covered in $10$ simulated seconds. The first ($x$-axis) dimension of the feature space is the percentage of voxels that are the stiff bone (dark blue) material. The second feature-space dimension is the percentage of voxels filled. The resolution of the map is 128 $\times$ 128. We launched 10 runs for each treatment, but some had not completed in time to be included in this draft of the paper. We thus include data only from runs that finished in our plots and statistical analyses (7 for the EA treatment, 5 for the EA+Diversity treatment, and 8 for the MAP-Elites treatment). In later drafts of this paper we will report on a complete set of finished experiments, which will also have a larger and consistent number of runs per treatment. 

Our two control algorithms are implemented in NSGA-II and have been used in previous studies \cite{Mouret2012,tarapore2015evolvability}: (1) a single-objective evolutionary algorithm optimizing performance only, which we refer to as the ``traditional EA'' or just ``EA'' for short, and (2) a two-objective evolutionary algorithm that optimizes performance and diversity, which we call EA+D. Diversity is measured for each individual as the average distance in the feature space to every other individual. Both control treatments performed the same number of evaluations as MAP-Elites.

\begin{figure*}[ht!]
     \begin{center}
     \subfigure{
            \label{fig:first}
            \includegraphics[width=\textwidth]{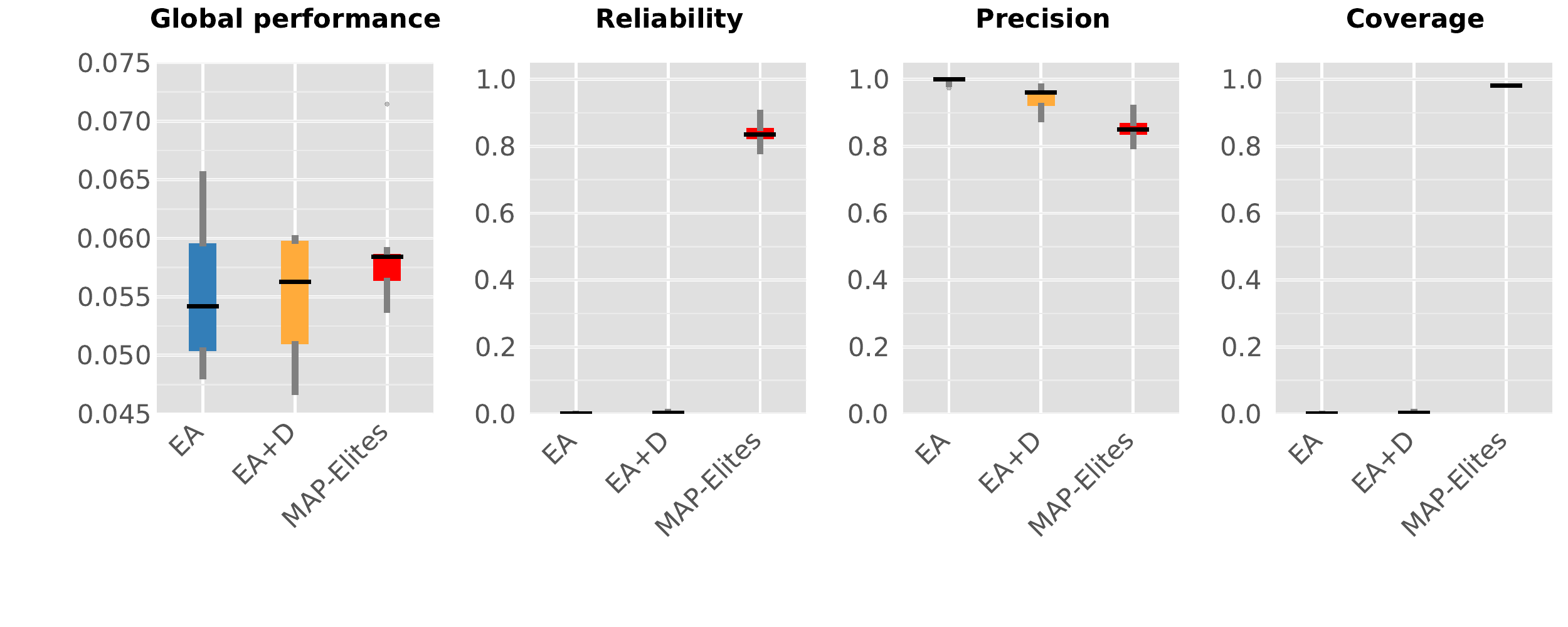}
        }\\
        \subfigure{
            \label{fig:first}
            \includegraphics[width=0.3\textwidth]{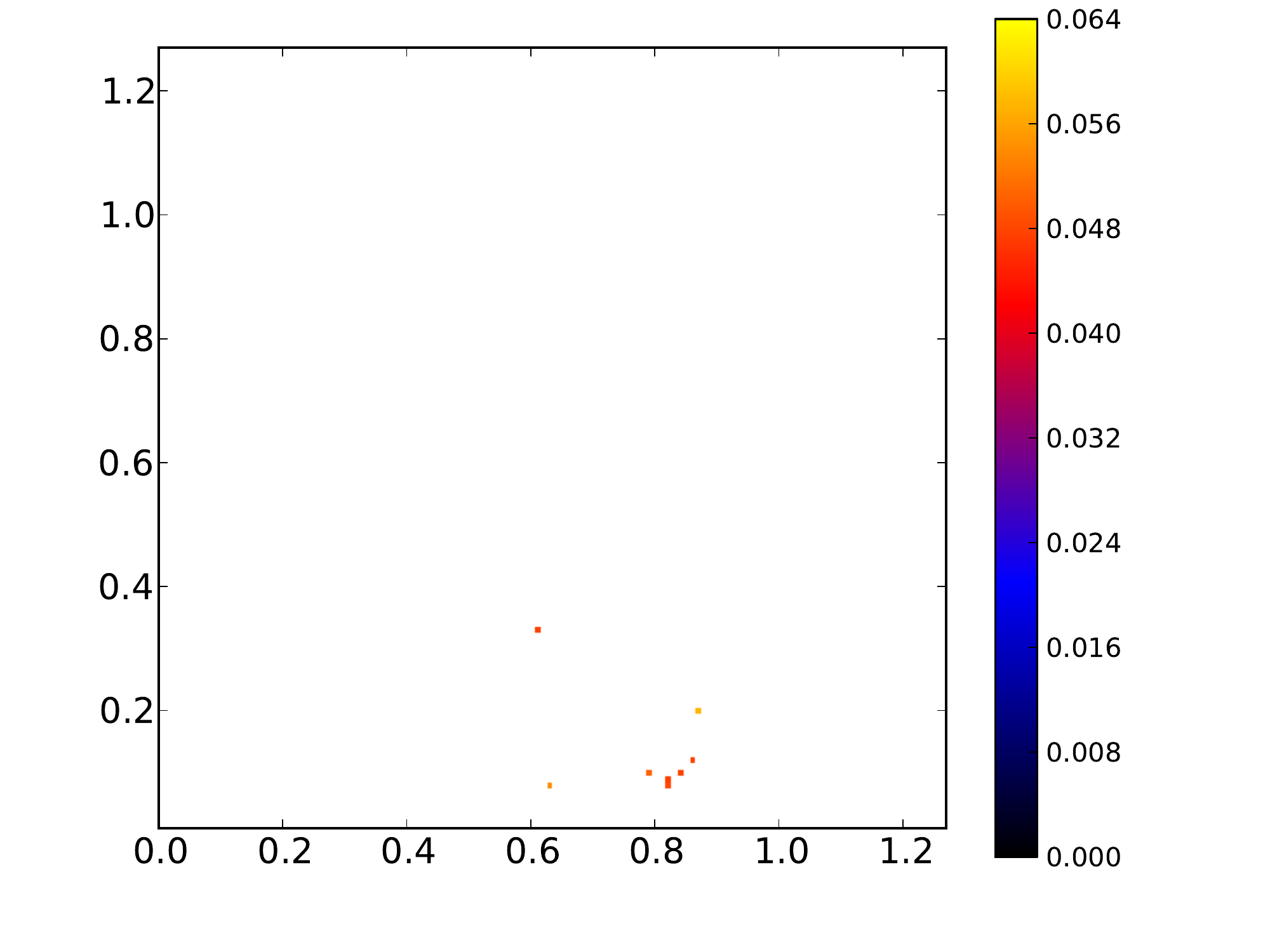}
        }
        \subfigure{
            \label{fig:first}
            \includegraphics[width=0.3\textwidth]{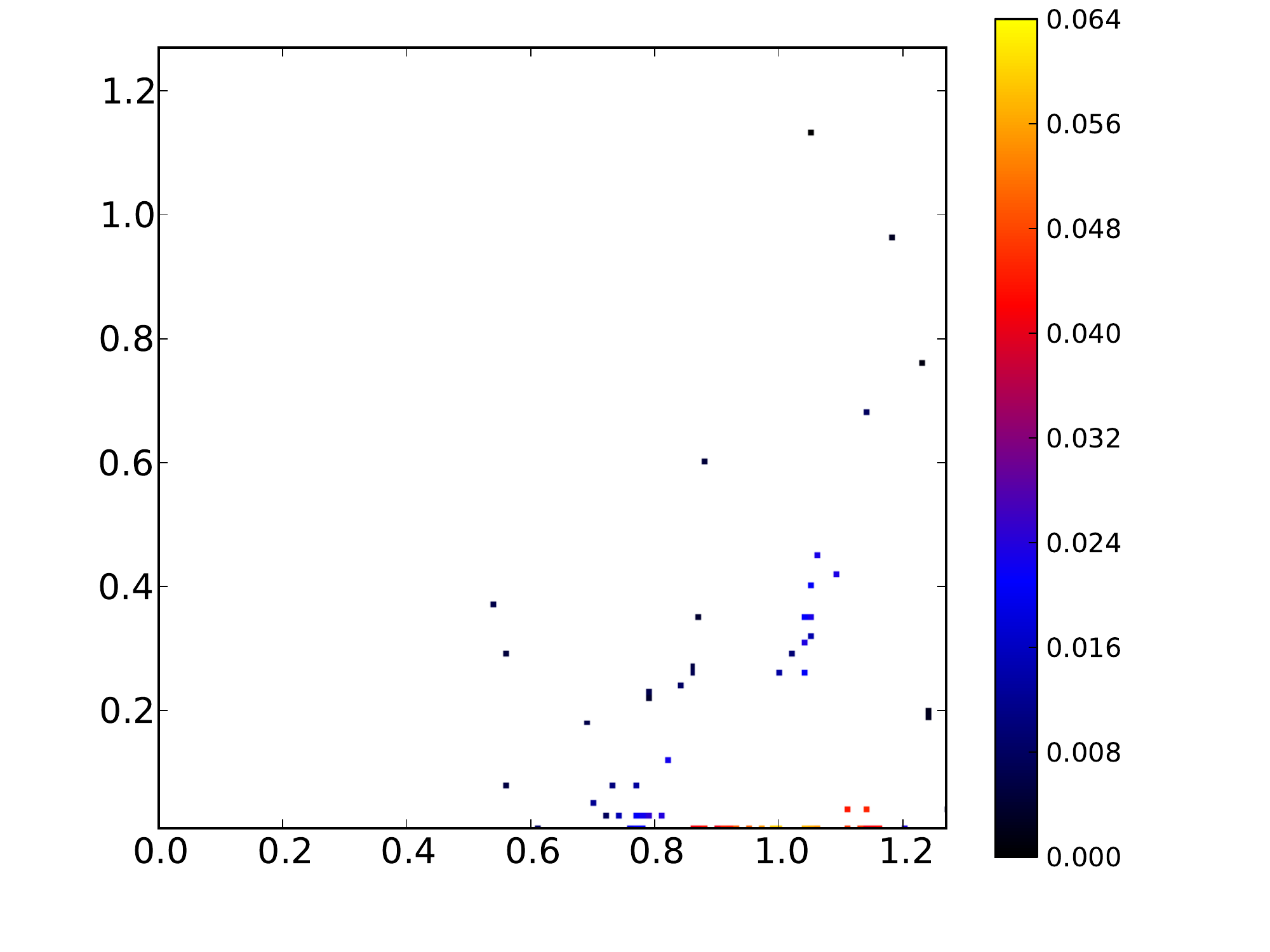}
        }
        \subfigure{
           \label{fig:first}
           \includegraphics[width=0.3\textwidth]{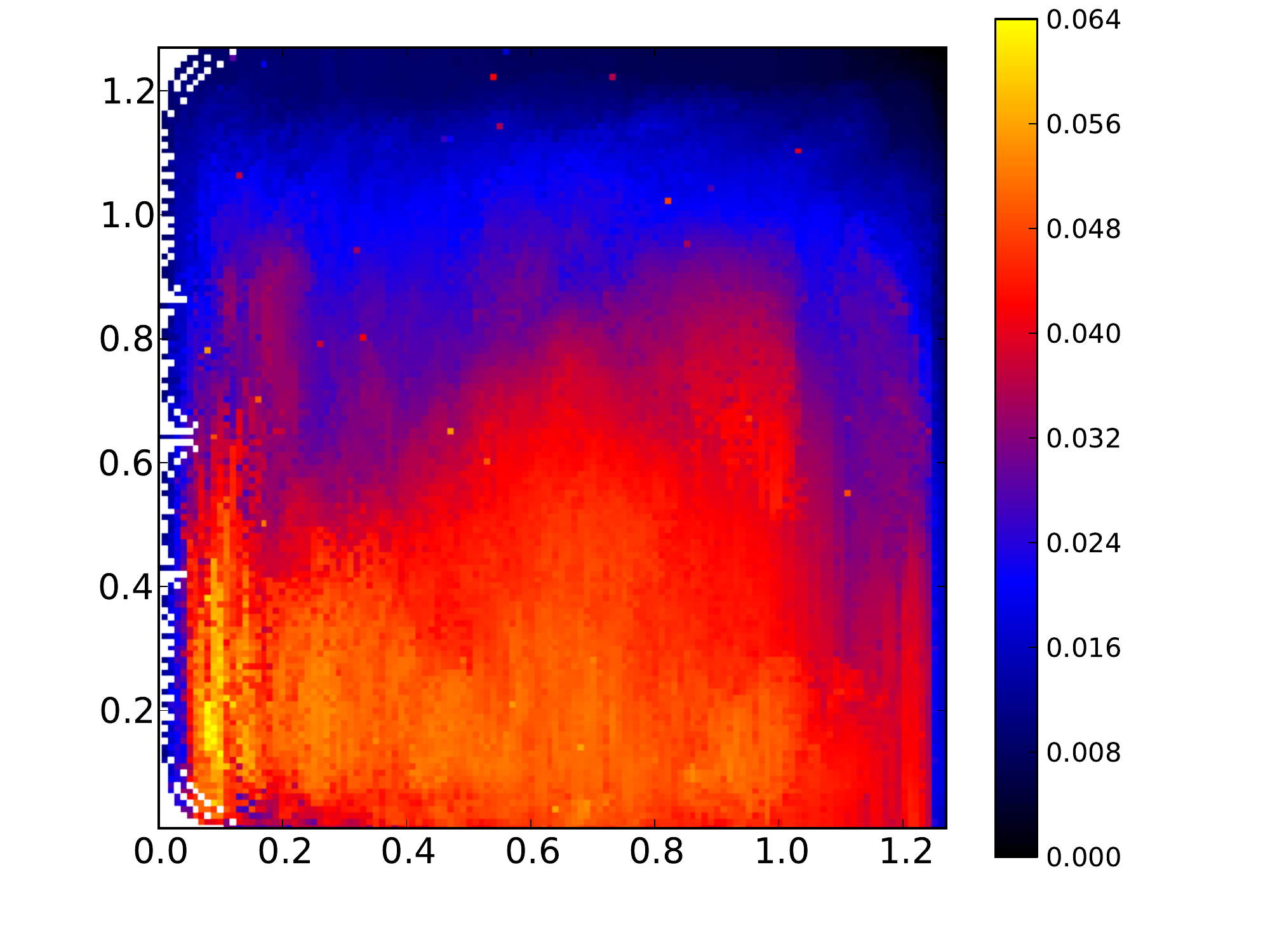}
        }\\
         \subfigure{
            \label{fig:first}
             \begin{minipage}{0.3\textwidth}
            \includegraphics[width=\textwidth]{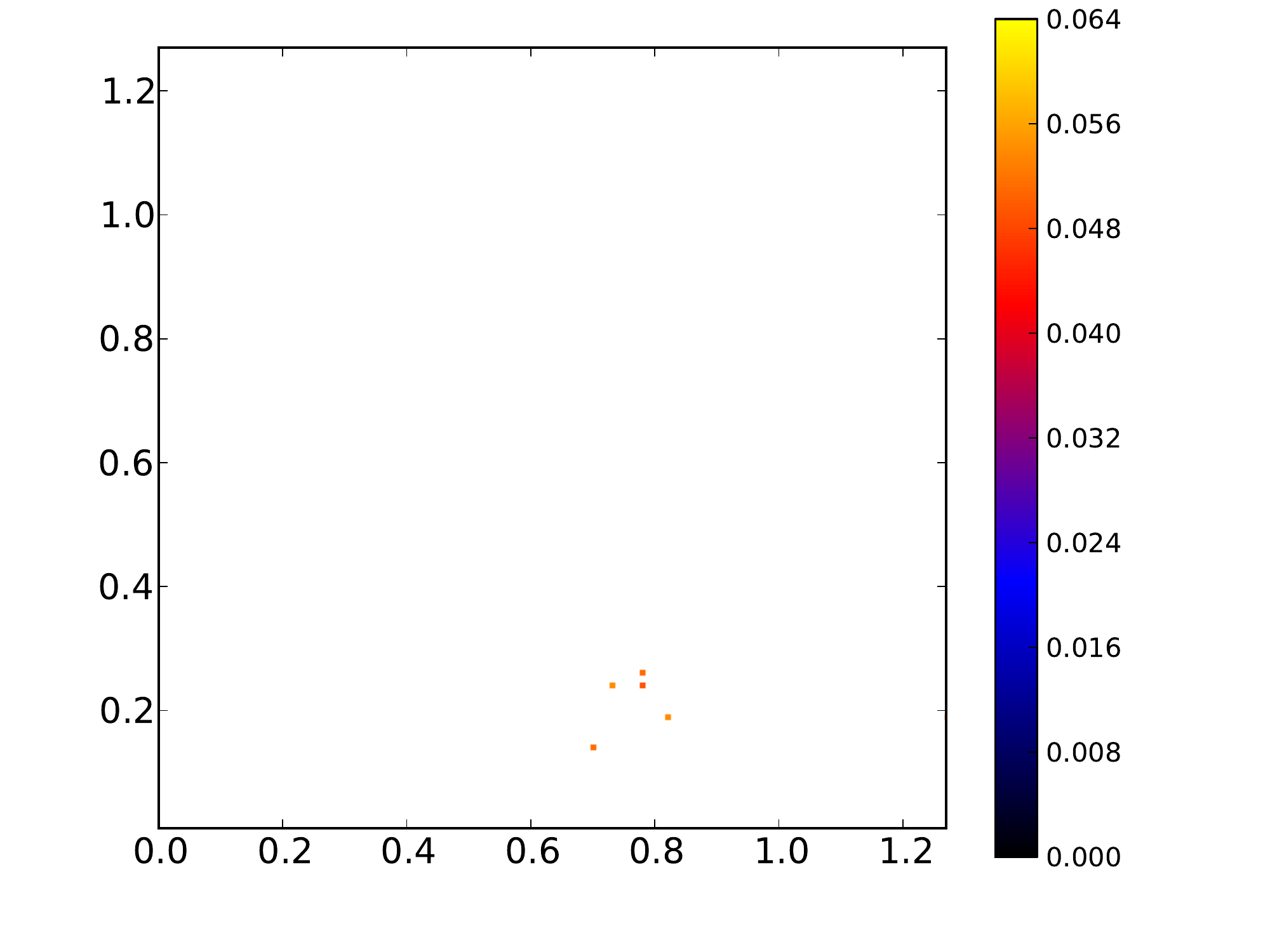}
            \begin{center}
            (a) EA
            \end{center}
            \end{minipage}
        }
          \subfigure{
            \label{fig:first}
            \begin{minipage}{0.3\textwidth}
            \includegraphics[width=\textwidth]{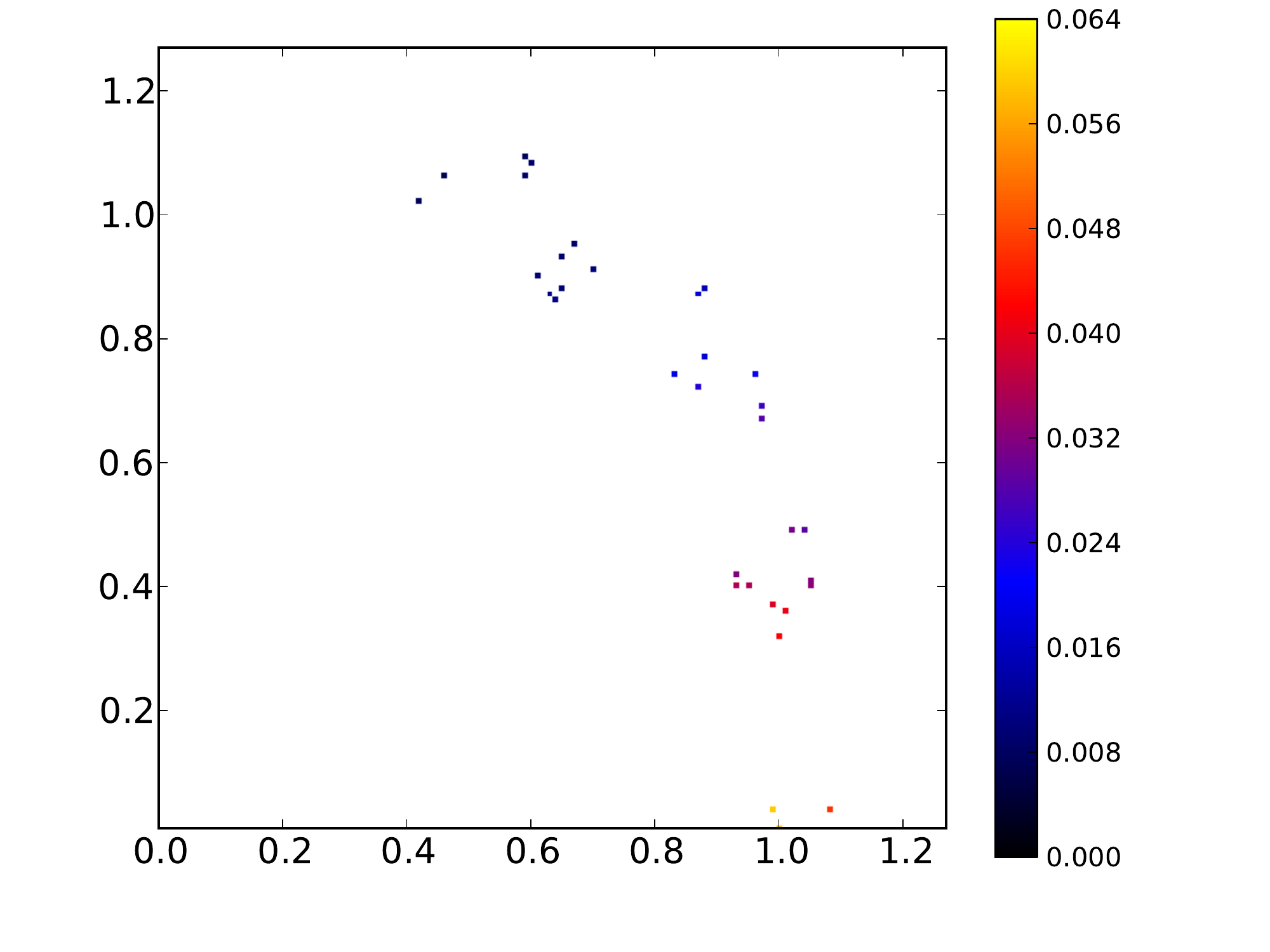}
            \begin{center}
            (b) EA + D
            \end{center}
            \end{minipage}
        }
         \subfigure{
            \label{fig:first}
            \begin{minipage}{0.3\textwidth}
            \includegraphics[width=\textwidth]{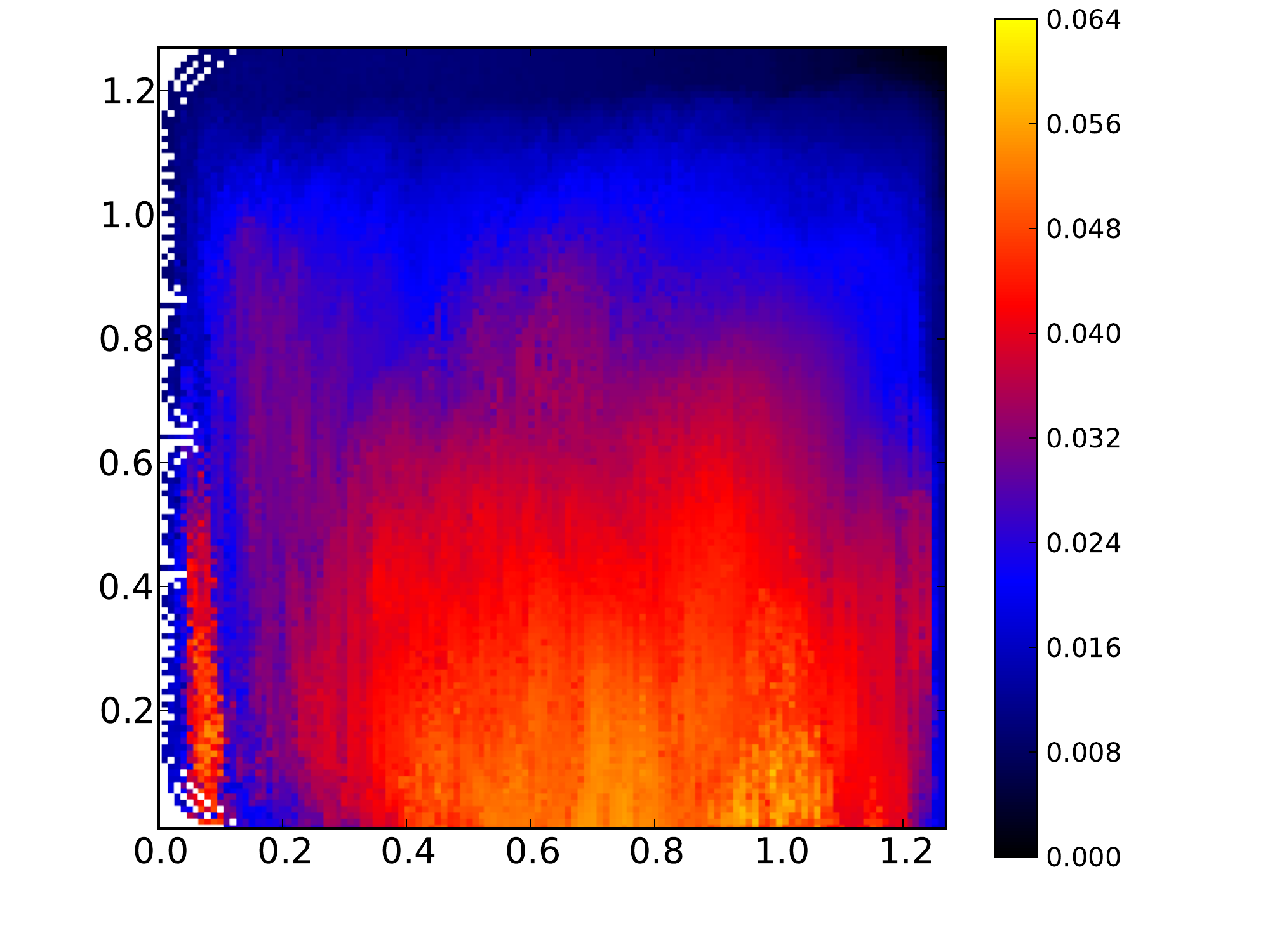}
            \begin{center}
         	   (c) MAP-Elites
         	   \end{center}
         	 \end{minipage}
        }         
    \end{center}
    \caption{%
        MAP-Elites does much better than a traditional evolutionary algorithm (EA) and an EA with diversity (EA+D) at finding high-performing solutions throughout a feature space. Data are from the simulated, soft robot morphologies problem domain. \textbf{Top:} MAP-Elites significantly outperforms the controls in global reliability and coverage (top, $p<0.002$). \textbf{Bottom:} Qualitatively, example maps produced by two independent runs demonstrate MAP-Elites' ability to both fill cells (coverage) and reveal the fitness potential of different areas of the feature space. Note the difference in feature-space exploration between the MAP-Elites illumination algorithm and the optimization algorithms. 
     }%
   \label{fig:softbotsPlots}

\end{figure*}

In this domain, MAP-Elites does a far better job than the controls of revealing the fitness potential of each area of the feature space, which is the goal of illumination algorithms~(Fig.~\ref{fig:softbotsPlots}). It has significantly higher reliability and coverage ($p<0.002$), and example maps highlight the tremendous difference in terms of exploring the feature space between the MAP-Elites illumination algorithm and the two control optimization algorithms, even though one has a diversity pressure.

In terms of global performance, while MAP-Elites has a higher median value, there is no significant difference between it and the other treatments ($p > 0.05$). If one cared only about finding a single, high-performing solution, then there would thus be no statistical difference between MAP-Elites and the two optimization algorithm controls. However, if one wanted a variety of different, high-performing solutions, MAP-Elites produces far more.   

MAP-Elites is significantly worse at precision than the two control algorithms ($p<0.01$). This result is likely explained by the fact that the control algorithms allocate all of their evaluations to very few cells, and thus find good solutions for those cells. In contrast, MAP-Elites has to distribute its evaluations across orders of magnitude more cells, making it hard to always find a high-performing solution in each cell. Note that MAP-Elites is usually competing against itself in this regard: because there is so little exploration by the control algorithms, they rarely produce the highest-performing solution across all runs of all treatments for a particular cell. Those instead tend to come from MAP-Elites.\todoOfficialVersion{This could be another measure: who produces the best ever for each cell} Thus, most low precision scores for MAP-Elites come when one run of MAP-Elites does not find as high-performing a solution in a cell as another run of MAP-Elites. We hypothesize that if each run of MAP-Elites were given more evaluations (i.e. run longer), it would catch up to, if not surpass, the controls in precision. That is a beneficial, and rare, property for an evolutionary algorithm to have: that it can benefit from additional computation because it does not get stuck on local optima and cease to innovate. 

\begin{figure*}[ht!]
	\begin{center}
        		\subfigure{
		\label{fig:softbotThumbnails1}
             	\begin{minipage}{0.89\textwidth}
	        \includegraphics[width=\textwidth]{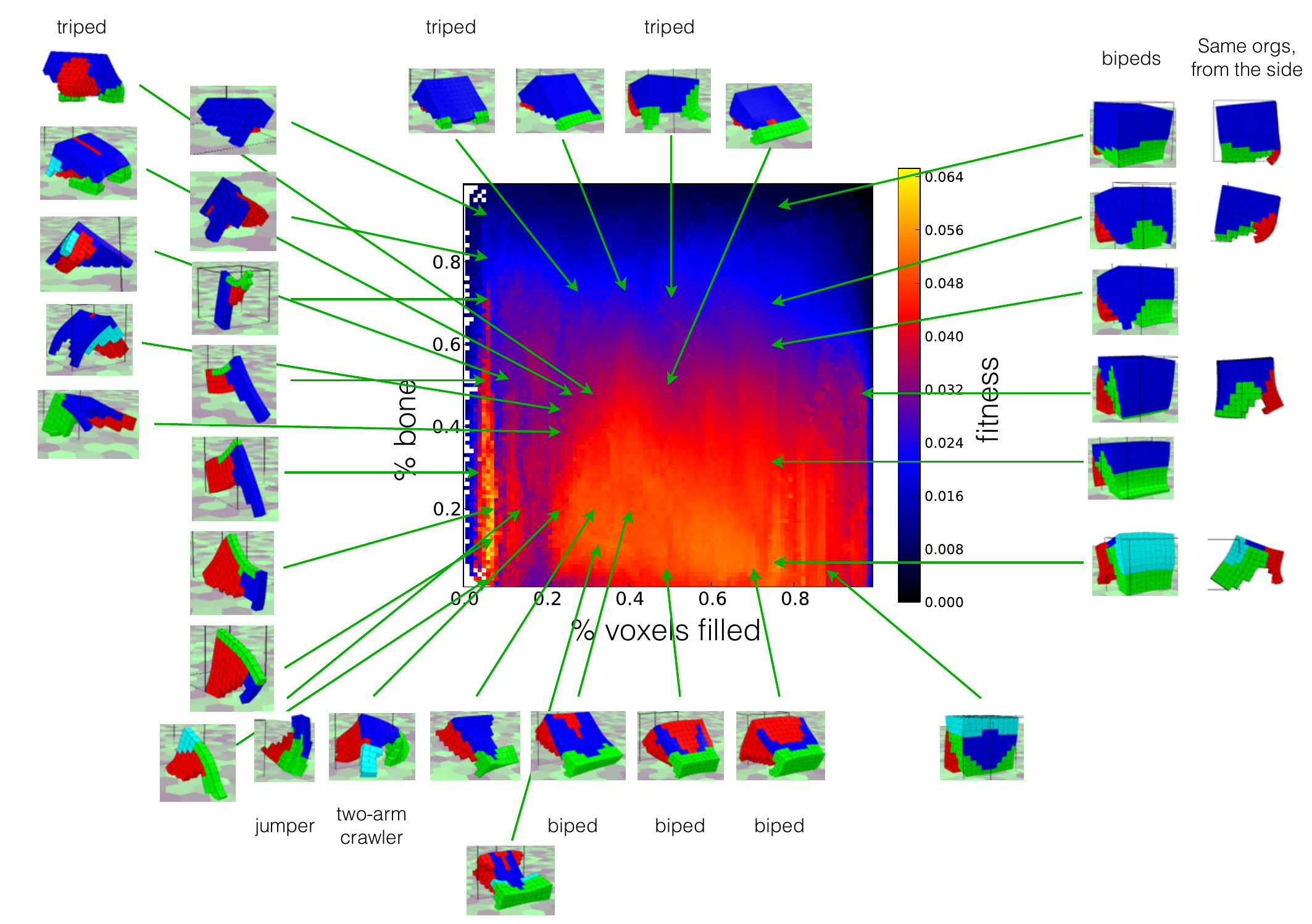}
            	\end{minipage}
        		}
        		\subfigure{
		\label{fig:softbotThumbnails2}
             	\begin{minipage}{0.89\textwidth}
	        \includegraphics[width=\textwidth]{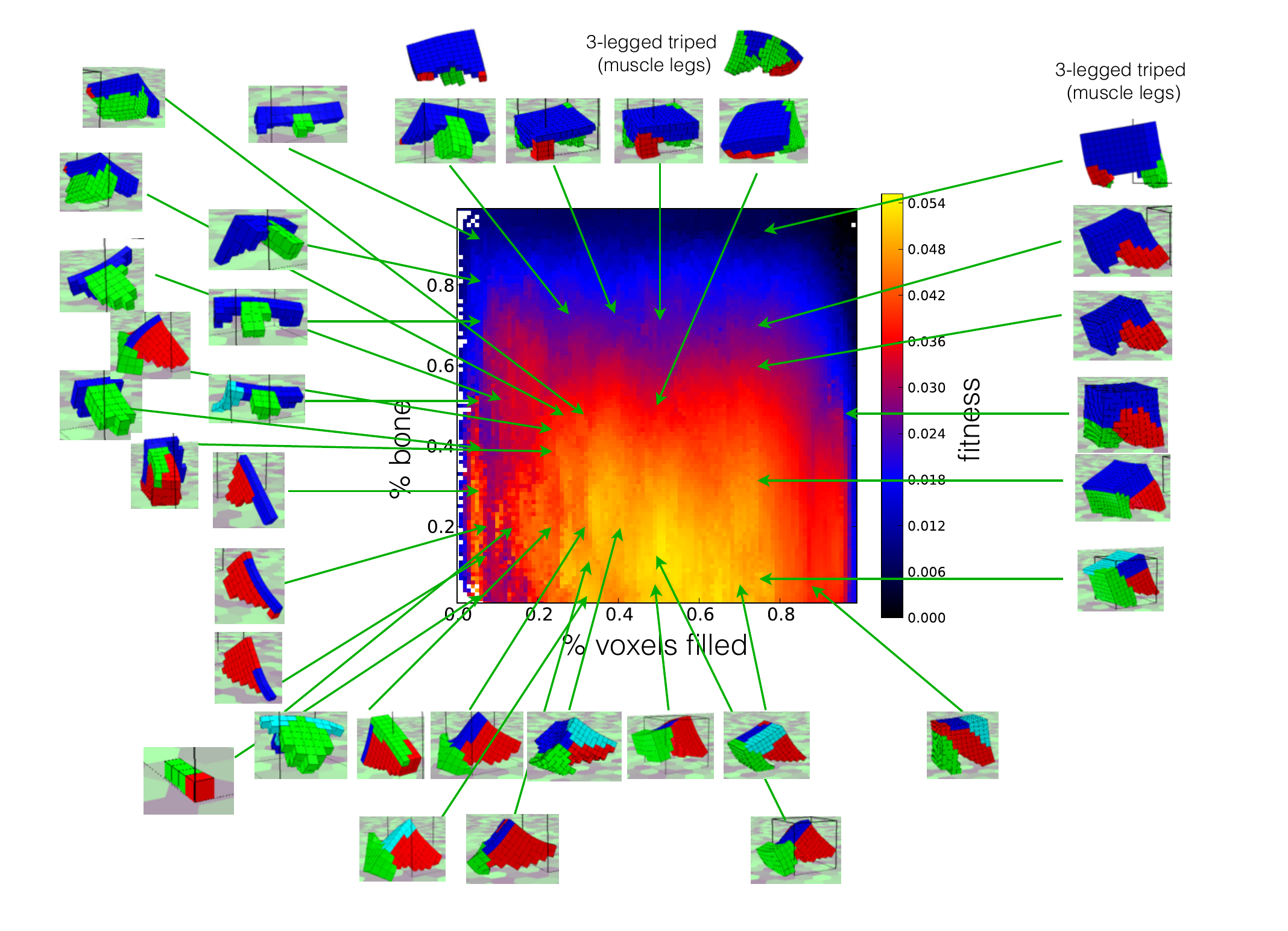}
            	\end{minipage}
        		}
	\end{center}
	\caption{Example maps annotated with example organisms from different areas of the feature space. Within a map, MAP-Elites smoothly adapts a design theme along the desired dimensions of variation. Between maps, one can see that there is some variation between maps, both in the performance discovered at specific points, and in the types of solutions discovered. That said, in general each map generally paints the same overall picture of the performance capabilities of each region of the feature space. Note the different scale of the bottom color map. Additional example maps are shown in Fig.~\ref{fig:softbotsPlots}. Because videos do a better job of revealing the similarity and differences in these organisms, both in their body and their behavior, a future draft of the paper will include a video of these individuals.}
	\label{fig:softbotThumbnails}
\end{figure*}

There is variance in the maps produced by independent runs of MAP-Elites. That reflects the fact that it is a stochastic search algorithm with historical contingency. The perfect illumination algorithm would always find the highest-performing solution at each point in the map, and thus have no between-run variance. However, while there are differences between the maps of different runs, they largely reveal the same overall pattern~(Figs.~\ref{fig:softbotsPlots} and \ref{fig:softbotThumbnails}).

By looking at pictures and videos of the elites in the final map of individual runs, we observed that MAP-Elites does indeed produce smooth changes across the chosen dimensions of variation (Fig.~\ref{fig:softbotThumbnails}). Consider the column of examples from the right side of Fig.~\ref{fig:softbotThumbnails}, Top, where the percent of voxels filled is roughly 75\%. Starting at the bottom, with around 10\% bone, there is a design with a red muscle hind leg, green muscle front leg, and no bone in the back connecting these legs (instead there is light blue soft tissue). Sweeping up in that column, the percentage of bone is increased, predominantly in the back connecting the legs, and the soft tissue and amount of muscle in each leg is reduced to gradually increase the amount of dark blue bone. These same creatures visualized from the side (rightmost column of images in Fig. \ref{fig:softbotThumbnails}, Top) shows that the basic biped locomotion pattern is preserved despite these changes, going from (in the bottom) a fast, flexible biped that resembles a rabbit, to a slow biped creature that resembles a turtle with a massive, heavy shell. MAP-Elites is thus achieving its goal of providing the requested variation and producing a high-performing solution of each type. True to the motivation of illumination algorithms~\cite{lehman2011evolving, clune2013originModularity}, finding the fastest, heavy shelled turtle does not preclude finding the fastest rabbit: in this case, they both can win the race. 

From these maps, one can also learn about the fitness potential of different regions of the feature space. For example, the previous example showed that, holding the percentage of voxels filled at 75\%, the more bone in a creature, the slower it is. The full map reveals that is generally true for almost all body sizes. The maps also reveal an interesting, anomalous, long, skinny island of high performance in the column where the percentage of voxels filled is roughly 7\%. It turns out that column contains a variety of different solutions that are all one voxel wide. Some quirk of the simulator allows these vertical sheet organisms to perform better than creatures that are more than one voxel wide. It might take hundreds or thousands of runs with traditional optimization algorithms to learn of this high-performing region of the space, but with MAP-Elites it jumps out visually in each map. Even within this island, we can still see smooth gradients in the desired dimensions of variation, starting with sheets made entirely of muscle and transitioning to sheets made mostly of bone. Space constraints prevent showing all of the final elites, but we consistently observed that one can start in nearly any location of the map and smoothly vary the designs found there in any direction. A second example map is provided (Fig.~\ref{fig:softbotThumbnails}, Bottom) to show that these findings are not limited to one run of MAP-Elites, but are consistently found in each map: while the actual design themes are different from map to map, the fact that MAP-Elites provides smooth changes in these themes according to the desired dimensions of variations is consistent.

\subsection{Real soft robot arm}

While the previous section featured simulated soft robots, in this section we test whether MAP-Elites can help find controllers for a real, physical, soft robot arm. The physics of this arm are quite complicated and difficult to simulate, making it necessary to perform all evaluations on the real robot.  That limits the number of evaluations that can be performed, requiring a small feature space. This domain thus demonstrates that MAP-Elites is effective even on a challenging, real-world problem with expensive evaluations and a small feature space. 

\begin{figure*}[ht!]
     \begin{center}
        \subfigure{ %
            \label{fig:first}
            \includegraphics[width=\textwidth]{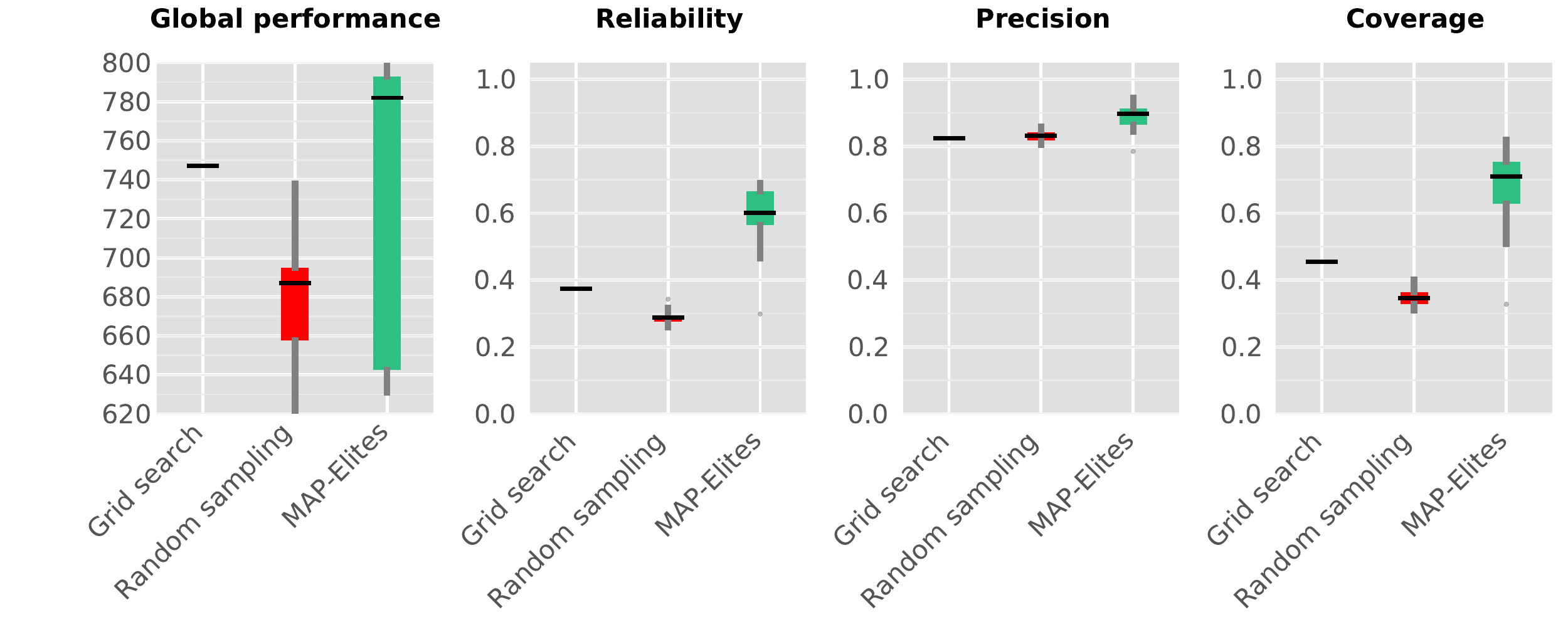}
        }
        \subfigure{
           \label{fig:second}
           \includegraphics[width=0.5\textwidth]{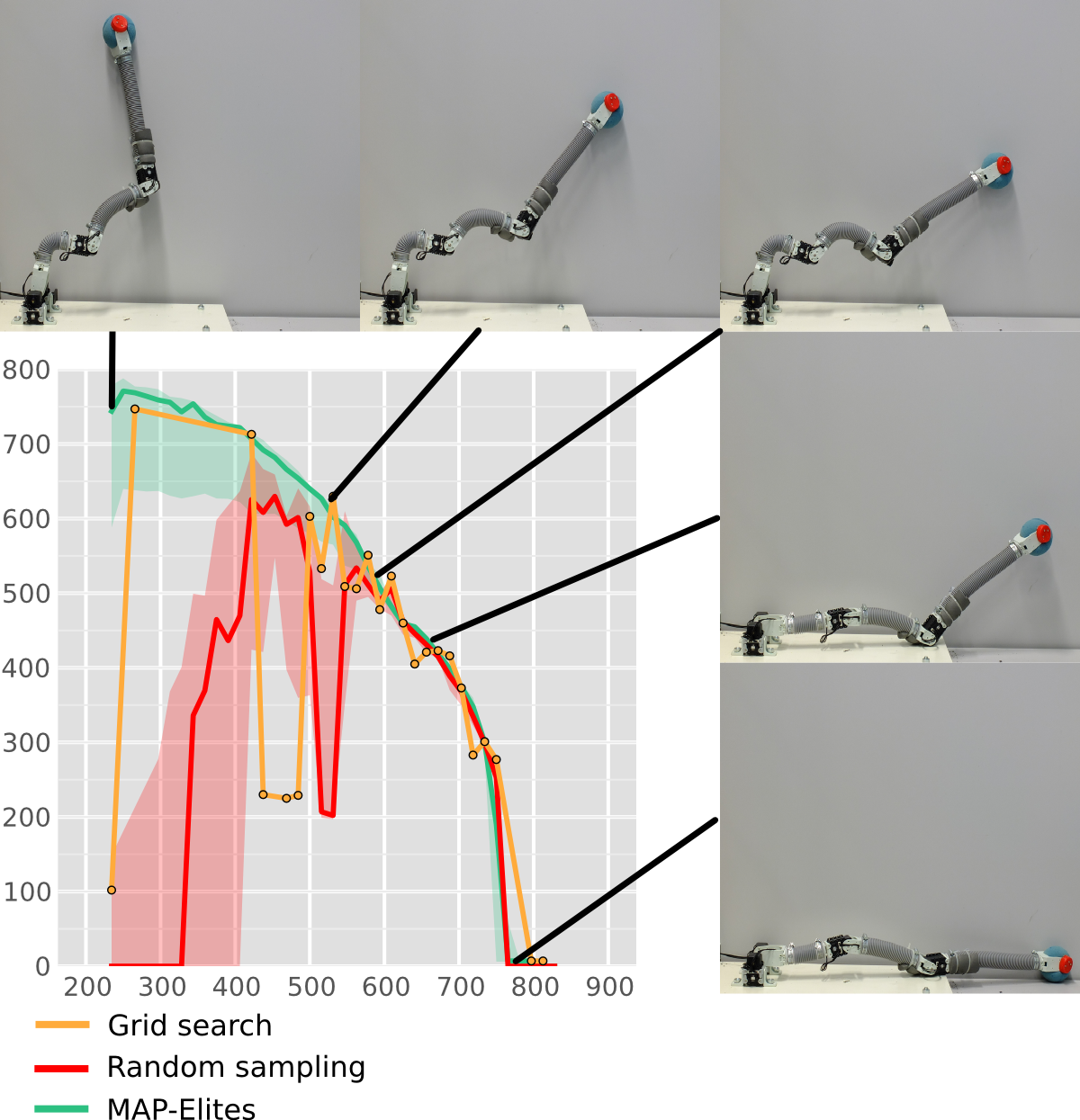}
        } 
    \end{center}
    \caption{%
    On a real, soft robot, MAP-Elites consistently finds high-performing solutions (higher $y$ values) across the feature space (different $x$  values) than controls.  
        \todoOfficialVersion{Why are there no circles on the green or red line, like there are for orange. Assuming that MAP-Elites is higher resolution, mention that}
     }%
   \label{fig:softArm}
\end{figure*}

We built a soft robotic arm (Fig.~\ref{fig:softArm}) by connecting 3 actuated joints (dynamixel AX-18 servos) with highly compliant tubes (made of flexible, washing machine drain pipes). An external camera tracked a red point at the end of the arm. A solution is a set of 3 numbers specifying the angle of each of the 3 joints. Specifically, each servo can move between -150 and +150 steps (out of the possible range for AX-18s of -512 and + 512 steps, which covers all 360 degrees of rotation). When the arm is fully extended and horizontal, the first servo from the base is at position 150, and the other two are at position 0. 

The feature space is one-dimensional: the $x$-value of the red circle at the end of the arm (in the coordinates of the image from the perspective of the camera). It is discretized into 64 cells. The performance function is to maximize the $y$-value of the end of the arm. The experiments thus attempt to discover the boundaries of the workspace of the robot, which is hard to compute analytically with a flexible robot.

We evaluated MAP-Elites and two controls:  random sampling and a traditional grid search algorithm. In the random sampling control, each solution is determined by randomly choosing an angle for each joint in the allowable range. The grid search algorithm specifies, for each joint, eight points evenly distributed within the range of allowable angles for that joint, and then tests all combinations of the possible values for each joint. We replicated each experiment 10 times, except for grid search, which is a deterministic algorithm and thus need only be run once. Each MAP-Elites and random sampling experiment were allocated 640 evaluations; the grid search required 729 evaluations ($9\times9\times9$).

The results show that all three treatments find approximately the correct boundary for high values of $x$ ({\raise.17ex\hbox{$\scriptstyle\sim$}}600 to {\raise.17ex\hbox{$\scriptstyle\sim$}}800). Our observations of the robot revealed why this is a relatively easy task: these points can be reached by setting the angles of the first and second joints (counting from the base) to put the third joint on the ground, and only changing the angle of this third joint. Because of the flexibility of the links, many different combinations of angles for the first and second joint result in having the wrist on the table.

Intermediate values of $x$ (approximately 400-600) represent harder problems, because in this range there are fewer joint angle values that combine to reach near the maximum height. MAP-Elites outperforms both grid search and random sampling in this region. Even when they are at their best, MAP-Elites tends to outperform these algorithms. For each control algorithm, there are values in this region were their performance is especially poor. 

Lower values of $x$ represent even harder challenges. Grid search found only a few points below 500, and thus provides a less-informative (lower-resolution) picture than MAP-Elites does. 
To have good coverage of this low-$x$-value region (200-500), we would need to significantly increase the resolution (discretization) of grid search, which would require exponentially more evaluations. The rarity of high-performing solutions in this part of the feature space results in even lower performance for random sampling. MAP-Elites, in contrast, provides many high-performing solutions for all values of $x$. These data are still too preliminary to provide reliable statistical results, but we plot them to show what we know to date about MAP-Elites and the controls in this problem domain. 

\section{Discussion and Conclusion}

This paper introduces the term ``illumination algorithms'' for the class of algorithms that try to find the highest-performing solution at each point in a user-defined feature space. It also introduces a new illumination algorithm called MAP-Elites, which is simpler to implement and understand than previous illumination algorithms, namely Novelty Search + Local Competition\cite{lehman2011evolving} and MOLE\cite{clune2013originModularity}.  Finally, the paper presents preliminary evidence showing that MAP-Elites tends to perform significantly better than control algorithms, either illumination algorithms or optimization algorithms, on three different problem domains. Because of the preliminary nature of the experimental data, we do not wish for readers at this point to conclude anything for certain yet about MAP-Elites' empirical performance, but in many cases the data  suggest that MAP-Elites is a promising new illumination algorithm that outperforms previous ones. 

Perhaps the best way to understand the benefits of illumination algorithms versus optimization algorithms is to view the feature maps from the simulated soft robot domain~(Fig.~\ref{fig:softbotsPlots}). Optimization algorithms may return a high performing solution, but they do not teach us about how key features of a search space relate to performance. MAP-Elites and other illumination algorithms, in contrast, map the entire feature space to inform users about what is possible and the various tradeoffs between features and performance. Such phenotype-fitness maps, as they are known in biology\cite{bull2011phenotype}, are interesting in their own right, and can also be put to practical use. For example, a recent paper showed that the map can provide a repertoire of different, high-performing solution that can initiate search for new behaviors in case an original behavior no longer works (e.g. if a robot becomes damaged or finds itself in a new environment)\cite{cully2015robots}.

MAP-Elites is also a powerful optimization algorithm, putting aside its additional benefits regarding illuminating feature landscapes. It significantly outperformed or matched control algorithms according to the narrow question of which algorithm found the single, highest-performing solution in each run. As discussed above, that could be because simultaneously searching for a multitude of different, related stepping stones may be a much better way to reach any individual stepping stone than directly searching only for a solution to that stepping stone\cite{nguyen2015introducing}. 

For a similar reason, illumination algorithms like MAP-Elites may help evolutionary algorithms move closer to the open-ended evolution seen in the natural world, which produced a tremendous diversity of organisms (within one run). In nature, there are a multitude of different niches, and a species being good in one niche does not preclude a different species from being good in another: i.e., that bears are stronger does not crowd out the ability for butterflies to flourish in their own way\cite{lehman2011evolving}. By simultaneously rewarding a multitude of different types of creatures, MAP-Elites captures some of that diversity-creating force of nature. One drawback to MAP-Elites, however, is that it does not allow the addition of new types of cells over time that did not exist in the original feature space. It thus, by definition, cannot exhibit open-ended evolution. Nature, in contrast, creates new niches while filling others (e.g. beavers create new types of environments that other species can specialize on). Future work is required to explore how to create illumination algorithms that do not just reveal the fitness potential of a predefined feature space, but that report the highest-performing solutions at each point in an ever expanding feature space that is intelligently enlarged over time.

In conclusion, illumination algorithms, which find the highest-performing solution at each point in a user-defined feature space, are valuable new tools to help us learn about complex search spaces. They illuminate the fitness potential of different combinations of features of interest, and they can also serve as powerful optimization algorithms. MAP-Elites represents a simple, intuitive, new, promising illumination algorithm that can serve these goals. It also captures some of the diversity generating power of nature because it simultaneously rewards the highest-performing solutions in a multitude of different niches. 

\section{Alternate variants of MAP-Elites}
\label{alternateVersions}

The following are alternate ways to implement MAP-Elites. Future research is necessary to see whether, and on which types of problems, any of these variants is consistently better than the simple, default version of MAP-Elites used in this paper.

Possible variants of this algorithm include:
\begin{itemize}

\item{Storing more than one genome per feature cell to promote diversity}

\item{Biasing the choice of which cells produce offspring, such as biasing towards cells who have empty adjacent cells, cells near low-performing areas, cells near high-performing areas, etc.} In preliminary experiments, such biases did not perform better than the default MAP-Elites.

\item{Including crossover. Crossover may be especially effective when restricted to occurring between organisms nearby in the feature space. Doing so allows different competing conventions in the population (e.g. tall, skinny organisms being crossed over only with other tall, skinny organisms, and the same for short, fat organisms). One could make crossover only occur within a certain radius of an individual or as a probabilistic function of the distance between organisms (leading to overlapping crossover zones), or only within certain regions (more akin to an island model). Note that even with geographically restricted crossover, offspring could still end up in different areas of the feature space than their parents (either due to the effects of mutation or crossover).}

\end{itemize}

\section{Methods}

\subsection{Statistics}

The statistical test for all $p$ values is a two-tailed Mann-Whitney U test. 

\subsection{Hierarchical, Parallelized MAP-Elites}
\label{hierarchicalVersion}

To first encourage a course-grained search, and then allow for increased granularity, we created a hierarchical version of MAP-Elites. It starts with larger cells and then subdivides those cells over time. 
In this hierarchical version of MAP-Elites, the sizes of cells shrink during search, and thus the range of differences in features that an organism competes with changes, although it is bounded to within a cell: competition is thus still restricted to solutions with similar features. 

To make MAP-Elites run faster on a supercomputer containing many networked computers, we created a batched, parallelized version of hierarchical MAP-Elites. It farms out batches of evaluations to slave nodes and receives performance scores and behavioral descriptors back from these nodes. These optimizations should not have any qualitative effect on the overall performance of the algorithm. All of the experiments in this paper were conducted with this hierarchical, paralleled version of MAP-Elites.

\subsection{Experimental parameters}

\paragraph{Retina experiments.}
20 replicates for each treatment.

The MAP-Elites parameters are as follows:
\begin{itemize}
\item starting size of the map: 16 $\times$ 16
\item final size of the map: 512 $\times$ 512
\item batch size: 2,000
\item number of iterations: 10,000
\item initial batch: 20,000
\item resolution change program (4 changes):
\begin{itemize} %
\item iteration 0: 64 $\times$ 64
\item iteration 1250: 128 $\times$ 128
\item iteration 2500: 256 $\times$ 256
\item iteration 5000: 512 $\times$ 512
\end{itemize}
\item feature 1: connection cost (see \cite{clune2013originModularity})
\item feature 2: network modularity (see \cite{clune2013originModularity})
\item performance: percent answers correct on retina problem\cite{clune2013originModularity}
\end{itemize}

\paragraph{Soft robots experiments.}

The MAP-Elites parameters are as follows:
\begin{itemize}
\item feature 1: percentage of bones
\item feature 2: percentage of voxels filled
\item performance: covered distance
\item starting resolution: 64 $\times$ 64
\item final resolution: 128 $\times$ 128
\item batch size: 1024
\item initial batch: 4096
\item iterations: 1400 %
\end{itemize}

\paragraph{Soft physical arm.}
10 replicates for each treatment except for the grid search, which is deterministic and thus requires only one run.

For the grid search:
\begin{itemize}
\item total number of evaluations: 512
\item discretization of the parameters: 8 steps
\end{itemize}

For the random sampling:
\begin{itemize}
\item total number of evaluations: 420
\item we report the best solution found in each of the 64 cells used by MAP-Elites
\end{itemize}

For MAP-Elites:
\begin{itemize}
\item total number of evaluations: 420
\item feature 1: $x$ coordinate
\item fitness: maximize height ($y$ coordinate)
\item starting resolution: 64
\item final resolution:  64
\item batch size: 10
\item initial batch: 120
\item iterations: 30
\end{itemize}

\subsection{Quantifiable measurements of algorithm quality}

The notation in this section assumes a two-dimensional feature map ($x$ and $y$), but can be generalized to any number of dimensions.

\subsubsection{Global reliability} 
\label{methodsGlobalReliability}

Measures how close the highest performing solution found by the algorithm for each cell in the map is to the highest possible performance for that cell, averaged over all cells in the map. Because we do not know the highest performance possible for each cell, we approximate it by setting it equal to the highest performance found for that cell by any run of any algorithm. Cells that have never been filled by any algorithm are ignored. If an algorithm in a run does not produce a solution in a cell, the performance for that algorithm for that cell is set to 0 because the algorithm found zero percent of that cell's potential. 

We first define $M_{x,y}$ as the best solution found across all runs of all treatments at coordinates $x, y$. If $\mathcal{M}=m_1, \cdots, m_k$ is a vector containing the final map from every run of every treatment, then
\begin{displaymath}
M_{x,y} = \max_{i \in [1, \cdots, k]} m_i(x, y)
\end{displaymath}
We then define the global reliability $G(m)$ of a map $m$ as follows:
\begin{displaymath}
G(m) = \frac{1}{n(M)} \sum_{x,y} \frac{m(x, y)}{M(x,y)}
\end{displaymath}
where $x, y \in \left\{[x_{min}, \cdots, x_{max};y_{min}, \cdots, y_{max}] \right\}$, and $n(M)$ is the number of non-zero entries in $M$ (i.e. the number of unique cells that were filled by any run from any treatment).

\subsubsection{Precision (opt-in reliability)} 
\label{methodsPrecision}

Same as global reliability, but for each run, the normalized performance is averaged only for the cells that were filled by that algorithm in that run. This measure addresses the following question: when a cell is filled, how high-performing is the solution relative to what is possible for that cell? 

Mathematically, the opt-in reliability, or precision, $P(m)$ of a map $m$ is:
\begin{displaymath}
P(m) = \frac{1}{n(m)} \sum_{x,y} \frac{m(x, y)}{M(x,y)}
\end{displaymath}
for $x, y \in \left\{[x_{min}, \cdots, x_{max};y_{min}, \cdots, y_{max}] | \textrm{filled}_{m}(x, y) = 1\right\}$, where 
$\textrm{filled}_{m}(x, y)$ 
is a binary matrix that has a 1 in an $(x,y)$ cell if the algorithm produced a solution in that cell and 0 otherwise,
and where $n(M)$ is the number of non-zero entries in $M$ (i.e. the number of unique cells that were filled by any run from any treatment).

\subsubsection{Coverage} 
\label{methodsCoverage}
For a map $m$ produced by one run of one algorithm, we count the number of non-empty (i.e. filled) cells in that map and divide by the total number of cells that theoretically could be filled given the domain (i.e. for which a genome exists in the search space that maps to that feature-space cell).  Unfortunately, we do not know this total number of cells that theoretically could be filled for the experimental domains in this paper. We approximate this number by counting the number of unique cells that have been filled by any run from any treatment. Using the notation of the previous two sections, this number is
$n(F_{M})$, 
where $F_{M} = \textrm{filled}_{M}$.

\section{References}
\begin{small}
\sffamily
\bibliography{mapElites}
\end{small}

\section{Acknowledgements} Thanks to Roby Velez, Anh Nguyen, and Joost Huizinga for helpful discussions and comments on the manuscript. 

\end{document}